\PassOptionsToPackage{numbers}{natbib}

\documentclass[pre,aps,twocolumn,floatfix,10pt]{revtex4-2}

\usepackage{amsmath, amssymb}
\usepackage{graphicx}
\usepackage{dcolumn}
\usepackage{bm}
\usepackage{standard_setup}

\usepackage{soul}
\definecolor{lightskyblue}{rgb}{0.53, 0.81, 0.98}

\begin{document}


\title{Landscape Complexity for the Empirical Risk of Generalized Linear Models:\\
Discrimination between Structured Data}

\author{Theodoros G. Tsironis}
\author{Aris L. Moustakas}%
\affiliation{%
 Department of Physics, National Kapodistrian University of Athens, Athens, Greece \\
 and Athena Research Center / Archimedes Research Unit, Athens, Greece
}
\thanks{This work was partially supported by project MIS 5154714 of the National Recovery and Resilience Plan ``Greece 2.0'' funded by the European Union under the NextGenerationEU Program.}


\begin{abstract}
We use the Kac-Rice formula and results from random matrix theory to obtain the average number of critical points of a family of high-dimensional empirical loss functions, where the data are correlated $d$-dimensional Gaussian vectors, whose number has a fixed ratio with their dimension. The correlations are introduced to model the existence of structure in the data, as is common in current Machine-Learning systems. Under a technical hypothesis, our results are exact in the large-$d$ limit, and characterize the annealed landscape complexity, namely the logarithm of the expected number of critical points at a given value of the loss.

We first address in detail the landscape of the loss function of a single perceptron and then generalize it to the case where two competing data sets with different covariance matrices are present, with the perceptron seeking to discriminate between them. The latter model can be applied to understand the interplay between adversity and non-trivial data structure. For completeness, we also treat the case of a loss function used in training Generalized Linear Models in the presence of correlated input data. 
\end{abstract}


\maketitle


\section{Introduction} \label{sec:intro}

The celebrated success of Machine Learning (ML) is based on the effective use of mostly local iterative algorithms that are designed to ``learn'' features of data that are fed to them. Underlying the learning process is the efficient search of the local minima of cost (or loss) functions of the weights of nonlinear representations (such as perceptrons) of vast amounts of input data. 
Once the weights are determined they can be used, e.g. for discrimination purposes on datasets stemming from the same distribution. 
However, understanding the behavior of these algorithms is obstructed by the unsuitability of traditional statistical approaches in dealing with the fact that the size of the training sample, and the data dimension and weight dimensions are typically quite large and often of the same order \cite{lecun1998gradient,xiao2017fashion}.  
Furthermore, the performance of local algorithms, like gradient descent (or its variants), depends strongly on the geometry of the loss landscape they operate on, which is typically highly non-convex \cite{li2018visualizing}.
Therefore, the success on modern neural networks would suggest that although their loss-function might have many critical points (local minima and saddle-points), these points are typically rich in generalization. 
Nevertheless, on a fundamental level, the loss landscape of such systems is still poorly understood with only a handful of available results in this direction \cite{Choromanska_Henaff_Mathieu_Arous_LeCun_2015,Baskerville_Keating_Mezzadri_Najnudel_2022,Maillard_Arous_Biroli_2023,Arjevani_Field_2020,Arjevani_Field_2021,fyodorov2022optimization}. 

To provide more clarity in this issue, in this paper we will address the geometry of the loss function of a single perceptron, a simple model of a neuron, which was first introduced as a trainable classifier \cite{mcculloch43a}. 
It is also the building block of the first models used in statistical physics for devising a theory of learning \cite{Derrida_Gardner_Zippelius_1987,Gardner_Derrida_1988,Gardner_Derrida_1989} and has since been studied as a prototypical spin-glass model. 
In its simplest form, the loss function takes the form
\begin{equation} 
\label{eq:L}
    L(w) = \sum_{\mu=1}^m \sigma(w ^\intercal \xi_\mu)
\end{equation}
where $w\in\mathbb{S}^{d-1}$ is the weight vector and $\{\xi_\mu\}_{\mu=1}^m$ are a set of random data in $\mathbb{R}^d$.
In its traditional formulation, the activation function $\sigma$ is a step function with a specific threshold, but choosing a smooth activation function brings the model closer to modern machine learning considerations.
As a statistical physics model it has many interesting properties such as a jamming and glassy behavior \cite{franz2016simplest}.

We aim to provide qualitative information on these random loss landscapes by estimating the number of their critical points about each of their level sets. 
Specifically we are interested in the quantity 
\begin{equation}
    \mathcal{N}(B;L) = \# \{ L(w)\in B \ | \ \nabla L(w)=0\}
\end{equation}
in the non-trivial scaling limit where $d\to\infty$ and $m/d\to\beta>1$. The corresponding asymptotic annealed critical point complexity may be defined as
\begin{equation}\label{eq:complexity_def}
    \Psi(B,L) = \lim_{d\to\infty} \frac{1}{d}\log\mathbb{E}\mathcal{N}(B;L) 
\end{equation}
The expectation of $\mathcal{N}(B,L)$ over the random data vectors 
$\{\xi_\mu\}_{\mu=1}^m$ above can be obtained by applying the Kac-Rice formula \cite{azais2009level} 
\begin{equation}
\label{eq:kac-rice}
    \begin{split}
        \mathbb{E}\mathcal{N}(B,L) = \int_{\mathbb{S}^{d-1}} &dw \,\,p_{\nabla L(w)}(0) \  \cdot
        \\
        &\mathbb{E} \left[  \mathds{1}_{L(w)\in B}
      |\det\nabla^2L(w)| \Big| \nabla L(w)=0 \right] 
    \end{split}
\end{equation} 
where $p_{\nabla L(w)}$ is the probability density (with respect to the randomness of the data vectors) of $\nabla L(w)$. Although we limit our investigation to the expected number of critical points, we expect this number to be dominated by local minima at lower loss levels, and by saddle points with a macroscopic number of descent directions at higher loss levels. The reason is that, when conditioned on low enough values of $L(w)$, it will be exponentially rare for the Hessian $\nabla^2L(w)$ to be non-positive definite. 

Recent advances in random matrix theory have been applied to understand the expectation in \eqref{eq:kac-rice}, resulting to the first picture of the complex loss landscapes emerging from the interplay between randomness and high-dimensionality \cite{Auffinger,Choromanska_Henaff_Mathieu_Arous_LeCun_2015, Maillard_Arous_Biroli_2023,Ros_Fyodorov_2023,asgari2025localminimaempiricalrisk}.
But in practice, sources of complexity are not exhausted there. 
For example, further richness of the resulting loss landscapes stems from the structure of the data distribution, which is commonly far from white noise, contrary to the common theoretical assumption. 
This structure is exactly what algorithms seek to identify and leverage for discrimination or generation tasks. 
Each data vector can be generated from one of several distributions, corresponding to the different structure classes to be identified \cite{lecun1998gradient}.  
Alternatively, it can be generated from a distribution whose structure is to be learned and replicated \cite{kingma2013auto,goodfellow2014generative}
This structure is expressed, among other properties, in the correlations of the data distribution.
Specifically, even Gaussian distributions, when high-dimensional, can express surprisingly rich structure and capture highly complicated datasets \cite{seddik2020random}.
Furthermore, high-dimensional Gaussians are not far in terms of properties from Lipschitz functions of high-dimensional Gaussians \cite{boucheron2003concentration,wainwright2019high}, which in turn empirically seem to be able to approximate common datasets with high precision, as is evident from the success of modern generative models \cite{Sohl-Dickstein_Weiss_Maheswaranathan_Ganguli_2015, Song_Ermon_2019,Ho_Jain_Abbeel_2020}.

Another relevant factor of richness of loss landscapes, which is encountered in practice, is the existence of competition, i.e. the case where the models learn not only to accept data from a specific distribution, but also to reject data from other, related distributions.
In discriminating tasks this process is evident, as common datasets contain many different classes, some of which are similar and could confuse the model \cite{lecun1998gradient,xiao2017fashion}. 
In this context, an important example is the case of Generative Adversarial Networks (GANs) \cite{goodfellow2014generative}, where the discriminating model is trained to antagonize a generator, which learns to approximate the target distribution with increasing accuracy. 
The success of such models stems from the increasing competition between the two adversaries.

\section{Related Works} \label{sec:literature}

The study of the convergence properties of descent methods has a long history since their introduction \cite{cauchy1847methode}.
A significant achievement has been the 
proof of convergence of Stochastic Gradient Descent (SGD) in the convex setting \cite{robbins1971convergence}.
Later investigations have focused on the characterization of asymptotic trajectories and the limit points of SGD and narrowed down the locations of these limit points to a subset of the critical points \cite{benaim2006dynamics}. 
Generally, there has been given particular emphasis in the exclusion of saddle points as limiting points of local methods \cite{panageas2016gradient,panageas2019first,antonakopoulos2022adagrad}, since the loss landscape is designed with its minima in mind. 
These results are greatly illuminating of the mechanisms of SGD and its variants, but practical needs increasingly move away from the scenarios where they apply.
This is partly due to their inherent lack of any sort of non-asymptotic quantification of the rate of convergence, which becomes increasingly relevant as the relevant loss landscape become increasingly complex. 
Unfortunately, only a few results go beyond convexity and low dimensionality in a meaningful way for modern machine learning. 
These aspects have been studied, in limited, but relevant, models, in the physics literature \cite{saad1995exact} and have recently also been rigorously proven \cite{veiga2022phase}. 
Of course, such analyses can only be carried out when the loss landscape is fairly simple. 

Moving to the analysis of local algorithms on more relevant problems requires also understanding the geometry of increasingly complex landscapes.
A list of initial questions to be addressed in the investigation of any landscape include the calculation of the number of saddle points and the local minima.
In the physics literature there has already been an interest in answering these kinds of questions for random functions \cite{fyodorov2004complexity,fyodorov2005counting,fyodorov2007replica,fyodorov2014freezing}.
As a result, a toolkit drawing from statistical physics, random matrix theory and large deviations theory has been developed for computing the expected number of saddle points with specific properties \cite{Auffinger,Maillard_Arous_Biroli_2023}.
A crucial part of this toolkit can only provide conjectures, but significant progress has been made in developing rigorous methods and proving these results \cite{Arous_Bourgade_McKenna_2022}. 
In particular, this approach has been significantly enriched in the process of its application to understand complex landscapes that arise in the context of spin glasses \cite{Auffinger,Auffinger_Chen_2014,Baskerville_Keating_Mezzadri_Najnudel_2022,Fan_Mei_Montanari_2020,McKenna_2022,Ros_Fyodorov_2023,Ros_Ben_Arous_Biroli_Cammarota_2019}. 

The landscape analysis of spin glasses has provided insight that can be applied to machine learning problems \cite{Ros_Fyodorov_2023,Zdeborová_Krzakala_2016,Barbier_2020, Gamarnik_Moore_Zdeborová_2022}.
A number of more recent works have made more quantitative connections between spin glasses and machine learning, investigating the consequences that landscape complexity theory would imply for machine learning \cite{Choromanska_Henaff_Mathieu_Arous_LeCun_2015, Baskerville_Keating_Mezzadri_Najnudel_2022}.
These works are significant in that there is no other known approach for understanding the loss landscape of modern models in such a global manner, but they are also very limited in their domain of applicability. 
The limitations fall in one of two categories, either the models are taken directly from the field of spin glasses and are thus unconvincing in their correspondence with modern machine learning models \cite{Choromanska_LeCun_Arous_2015,Hartnett_Parker_Geist_2018}, or they only model a handful of neurons \cite{Maillard_Arous_Biroli_2023} and so distance themselves from the emergence properties of modern models.
Nevertheless, both cases contribute significantly in the understanding of the properties of loss landscapes, such as the randomness of the data sample, the scaling between data and parameters, the non-linearity and complex interactions between the weights.

The proposed models also suffer from limitations that can be somewhat mitigated by appropriate generalizations. 
One such limitation stems from ignoring the data structure, under the common assumption that the data are (potentially labeled) white noise \cite{Maillard_Arous_Biroli_2023}.
Overcoming this limitation could be achieved by the introduction of correlation for each datapoint. 
Another direction for generalization is the introduction of adversarial data, as was done for example in \cite{Baskerville_Keating_Mezzadri_Najnudel_2022}, in an attempt to model GAN loss-gain landscapes through spin glasses.
In the context of perceptron-based models, the adversarial data can be introduced as a second set of data, which the perceptron is motivated to be suppress. 

From an analytical point of view, the connection of these methods with random matrix theory emerges through a random matrix related to the Hessian of the loss function at each point of the parameter manifold.
Understanding the spectra of these random matrices is important to evaluate the number of critical points, whereas the analysis of the deviations of their largest eigenvalues from the bulk leads to estimations of the indices of critical points.

\section{Main Results} 
\label{sec:results}

In this work we consider the expected number of critical points of the loss function $L(w)$, $\mathbb{E}\mathcal{N}(U,L)$, as well as the expected number of critical point of a closely related family of loss functions, which appear in the training of Generalized Linear Models (GLMs). 
These models, in their planted version, aim to infer a vector $w_*$ from non-linear observations of the form $\sigma(w_*^\intercal \xi_\mu)$ over a dataset $\{\xi_\mu\}_{\mu=1}^m$. Such a loss function typically has the form
\begin{equation} \label{eq:L_1}
    L_1(w) = \frac{1}{m}\sum_{\mu=1}^m (\sigma(w ^\intercal \xi_\mu) - \sigma(w_* ^\intercal \xi_\mu))^2
\end{equation}
This task is seemingly much closer to those usually considered in machine learning, considering that one is trying to train a classifier on data $\{\xi_\mu\}_{\mu=1}^m$ to fit the labels $\{\sigma(w_*^\intercal\xi_\mu)\}_{\mu=1}^m$. 

The landscapes of these models have been studied in the case where the data are sampled from white noise, in the limit where the number of data scale with the data dimension so that $m/d\to\beta$ \cite{Maillard_Arous_Biroli_2023}.
In our attempt to understand how the structure of the data affects the loss landscape, we will remain in this asymptotic regime, but study the models in the case where the data distribution has non-trivial correlation structure, expressed through an arbitrary correlation matrix $\Sigma=R^\intercal R$, i.e. $\xi_\mu = Rz_\mu$ with $z_\mu$ being $d$-dimensional white noise, i.e. $z_\mu\sim N(0,I_d)$. In our study we will not fully investigate the necessary conditions on $\Sigma$ but instead limit our discussion to matrices with bounded spectral norm, well defined limiting spectral distribution and at most finitely many outlier eigenvalues, converging to specific values. 

Finally, we formulate a model based on the perceptron, that aims to optimize a loss function that also takes into consideration a dataset to be avoided. 
Suppose for example that $\sigma\in\mathbb{R}\to[0,1]$ and that we want to have an output as close to $0$ when the input comes from one dataset $\{\xi_\mu\}_{\mu=1}^{m_1}$ and as close to $1$ as possible, when the input originates from another dataset, namely $\{\xi_\mu\}_{\mu=m_1+1}^{m_1+m_2}$, thus discriminating between the landscapes. Then we might use the loss function
\begin{equation}
    \label{eq:L_2}
    \begin{split}
    L_2(w) &= \frac{1}{m_1}\sum_{\mu=1}^{m_1}\sigma(w^\intercal \xi_\mu^1)^2 + \frac{1}{m_2}\sum_{\mu=m_1+1}^{m_1+m_2}(1-\sigma(w^\intercal \xi_\mu^2))^2
    \\
    &= \frac{1}{m_1}\sum_{\mu=1}^{m_1}\sigma_1(w^\intercal R_1z_\mu) - \frac{1}{m_2}\sum_{\mu=m_1+1}^{m_1+m_2}\sigma_2(w^\intercal R_2z_\mu)
    \end{split}
\end{equation}
where, for reasons that will become apparent later, we have defined $\sigma_1$ and $\sigma_2$ in the obvious way.
As a result, we have two competing datasets, differentiated through the structure of their covariance matrices $G=R_1 R_1^\intercal$ and $\Sigma = R_2 R_2^\intercal$
Note that in the above loss function, as in the corresponding one in \eqref{eq:L}, there is no ``planted'', ``teacher'' direction. Here the relevant asymptotic limit is when the data scale with the dimension, defining as before the ratio $(m_1+m_2)/d\to\beta$. However, we also need both datasets to be macroscopically large in the sense that $(m_1+m_2)/m_1\to\alpha_1$ and $(m_1+m_2)/m_2\to\alpha_2=\frac{\alpha_1}{\alpha_1-1}$. 
It is important to keep in mind that the perceptron cannot meaningfully generalize in the discrimination task, since it only checks a one-dimensional projection of high-dimensional data, yet we study the resulting landscape as an illuminating scenario for the landscapes that result in the interplay between structure and competition. 

\subsection*{The Landscape Formula for $L(w)$} 

Since $\mathcal{N}(B,L)$ is typically an exponentially large quantity, the calculation of its mean will require the usage of Large Deviation Principles (LDPs) and our results will be expressing $\mathbb{E}\mathcal{N}(B,L)$ as an extremum of a Free Energy over an $\mathcal{O}(1)$ number of macroscopic order parameters. 
Starting with $L$, omitting explicit reference to the domains over which the extremizations of each parameter takes place, we have
\begin{equation} 
\label{eq:N}
    \begin{split}
        \Psi(B;L) &= \sup_{\rho,q,f,\kappa,\ell\in B} \bigg(  F^{SP}(\rho,q;\kappa,f) 
        \\ & \quad\quad 
        + \inf_{\nu} \lim_{\epsilon\to0^+} \text{extr}_{\phi,\bar{\phi}} 
         \bigg(F^{RM}(\phi,\bar{\phi};\rho,\kappa,\nu,\epsilon) 
        \\ & \qquad\qquad\qquad\qquad\qquad 
        + F^{S}(\ell,\kappa,f,\nu;\rho) \bigg) \bigg) 
    \end{split}
\end{equation} 
The common domain of $\rho$ and $q$ is the image of $\mathbb{S}^{d-1}$ under the mapping $w\to(w^\intercal\Sigma w, w^\intercal\Sigma^{-1}w)$. Whereas every component of \eqref{eq:N} depends on $\Sigma$ only through its limiting spectral distribution, the domain of $\rho$ and $q$ depends on the whole spectrum, including any outlier eigenvalues.
Furthermore, the domain of $f$ is $[\min_y \sigma'(y)^2, \max_y \sigma'(y)^2]$, while the domain of $\kappa$ is $[\min_y y\sigma'(y), \max_y y\sigma'(y)]$ and the domain of $\ell$ is $B$. 
The extremization over $\phi$ and $\bar{\phi}$ is unconstrained over the complex plane.
Finally, $\nu$ is the probability measure of a random variable, $\zeta$, and it is parametrized by three auxiliary, real, unconstrained parameters, $\bar{\kappa}$, $\bar{f}$ and $\bar{\ell}$.
The explicit dependence on these parameters can be seen through $\nu$'s density: 
\begin{equation}
    \nu(d\zeta) 
    = \frac{1}{Z(\bar{\ell},\bar{\kappa},\bar{f};\rho)} 
    e^{-\frac{1}{2}\zeta^2-\frac{1}{\beta}F_\nu(\zeta;\bar{\ell},\bar{\kappa},\bar{f},\rho)}
    d\zeta
\end{equation}
where
\begin{equation} 
\label{eq:data_bias}
    F_\nu(\zeta;\bar{\ell},\bar{\kappa},\bar{f},\rho) = \bar{\ell}\sigma(\sqrt{\rho}\zeta)
    +\bar{f}\sigma'(\sqrt{\rho}\zeta)^2
    +\bar{\kappa}\sqrt{\rho}\zeta\sigma'(\sqrt{\rho}\zeta)
\end{equation}
and
\begin{equation}
    Z(\bar{\ell},\bar{\kappa},\bar{f};\rho) = \int d\zeta 
    e^{-\frac{1}{2}\zeta^2-\frac{1}{\beta}F_\nu(\zeta;\bar{\ell},\bar{\kappa},\bar{f},\rho)}
\end{equation}
and for the above expectation to be defined for all values of the auxiliary barred parameters, we need $\sigma$ to be even or of subquadratic growth.

The normalization $Z$ is closely related to $F^{S}$, the part of \eqref{eq:N} that controls the deviation of the empirical distribution of the input of the perceptron function from the normal distribution. In particular, we have
\begin{multline}
\label{eq:sanov_and_lagrange}
    F^{S}(\ell,\kappa,f,\nu;\rho)  = \bar{\ell}\ell +\bar{\kappa}\kappa +\bar{f}f +  \beta\log Z(\bar{\ell},\bar{\kappa},\bar{f};\rho)
\end{multline}
It becomes obvious here that the role of the auxiliary barred parameters is to fix the values of the means of the random variables they multiply in \eqref{eq:data_bias} with respect to the distribution $\nu(\cdot)$.

$F^{SP}$ contains the probability that a generic point on the sphere is a saddle point and controls the deviations of the geometric parameters of the system, $\rho = w^\intercal \Sigma w$ and $q = w^\intercal \Sigma^{-1} w$
\begin{equation}
    \begin{split}
        F^{SP}(\rho,q;\kappa,f)  =& \frac{1}{2}\log(\beta e) -\frac{1}{2}\log(f)  -\frac{1}{2}\mathbb{E}_{\lambda_\Sigma}\log\left(\lambda_\Sigma\right) 
        \\ & \qquad \qquad
        +\frac{1}{2}\beta\frac{\kappa^2}{f\rho}\left(1 - \rho q \right)
        -I(\rho,q)
    \end{split}
\end{equation}
where $\lambda_\Sigma$ is a random variable distributed according to the limiting spectral distribution of $\Sigma$ and the rate function $I$ is given by
\begin{equation}
    I(\rho,q) = \sup_{\bar{\rho},\bar{q}}\frac{1}{2}\mathbb{E}_{\lambda_\Sigma}\log\left[
        (1-\bar{\rho}\rho -\bar{q}q)I
        +\bar{\rho}\lambda_\Sigma
        +\bar{q}\lambda_\Sigma^{-1}
        \right]
\end{equation}

Lastly, $F^{RM}$ controls the large deviations of the determinant of the Hessian of the loss function at each saddle point
\begin{equation}
\label{eq:F_RM}
    \begin{split}
        F^{RM}&(\phi,\bar{\phi};\rho,\kappa,\nu,\epsilon) = \beta\log\mathbb{E}_\zeta\left|1+\frac{1}{\beta} \sigma''(\sqrt{\rho}\zeta)\phi\right|
        \\ & \qquad\quad\quad
        +\mathbb{E}_{\lambda_\Sigma}\log\left|\left(\kappa + \kappa \bar{\phi}\lambda_\Sigma\right) \right|
        +\Re((\kappa+i\epsilon)\bar{\phi} \phi)
    \end{split}
\end{equation}

\subsection{The Landscape Formula for $L_1(w)$} 

Moving on to the loss function related to GLMs we have a similar formula for the complexity
\begin{equation} 
\label{eq:N_1}
    \begin{split}
        \Psi(B;L_1) = &\sup_{\rho,q,p,r,f,\kappa,\kappa',\ell\in B} \bigg(  
        F^{SP}_1(\rho,q,r,p;\kappa,\kappa',f) 
        \\ &
        + \inf_{\nu} \lim_{\epsilon\to0^+} \text{extr}_{\phi,\bar{\phi}}  
        \bigg( 
        F^{RM}_1(\phi,\bar{\phi};\rho,p,\kappa,\nu,\epsilon) 
        \\ & \qquad \qquad \qquad \quad 
        + F^{S}_1(\ell,\kappa,\kappa',f,\nu;\rho,p) \bigg) \bigg)
    \end{split}
\end{equation}
but there are some notable changes. Firstly, there are new functions of the input of the perceptron that must be controlled, changing the meaning of $\kappa$, $f$ and $\ell$ and introducing $\kappa'$. Additionally, we need to control a new pair of geometric variables, given by $p=w^\intercal \Sigma w_*$ and $r=w^\intercal w_*$. The expressions of the free energies that appear in \eqref{eq:N_1} are detailed in Section \ref{sec:glm} for brevity. 

\subsection*{The Landscape Formula for $L_2(w)$} 

Finally, the expected number of critical points and hence the complexity for the loss function in \eqref{eq:L_2} can be expressed compactly in the same form as \eqref{eq:N}, but the functions and the meaning of the underlying parameters change significantly
\begin{equation} 
\label{eq:N_2}
    \begin{split}
        \Psi(B;L_2) &= \sup_{\rho,q,f,\kappa,\ell\in B} \bigg(  F^{SP}_2(\rho,q;\kappa,f) 
        \\ & \quad\quad 
        + \inf_{\nu} \lim_{\epsilon\to0^+} \text{extr}_{\phi,\bar{\phi}} 
         \bigg(F^{RM}_2(\phi,\bar{\phi};\rho,\kappa,\nu,\epsilon) 
        \\ & \qquad\qquad\qquad\qquad\qquad 
        + F^{S}_2(\ell,\kappa,f,\nu;\rho) \bigg) \bigg) 
    \end{split}
\end{equation}
The most important difference is that the underlying variables $\kappa$, $f$, $\rho$, $\phi$, as well as their barred counterparts and $\epsilon$ are now 2-dimensional, e.g. $\kappa=(\kappa_1,\kappa_2)$ etc, containing one value for each dataset. Furthermore, $q$ is now a 4-dimensional variable, controlling various geometric properties of the system, mixing $\Sigma$ and $G$. More details can be found in Section \ref{sec:discriminator}.

\section{The Kac Rice Formula for the Perceptron with Correlated Data: $L(w)$ } \label{sec:perceptron}

We will start with details of the calculation for $\mathbb{E}\mathcal{N}(U;L)$ and then we will explain how they generalize for the other models. 
Our analysis follows in general terms the recipe presented in \cite{Maillard_Arous_Biroli_2023} for the perceptron with uncorrelated data, although the existence of the covariance matrices will result to significant changes in all the steps.
The starting point is the Kac-Rice formula \eqref{eq:kac-rice}, which we rewrite in the following form
\begin{equation}
\label{eq:kac-rice_*}
    \begin{split}
        \mathbb{E}\mathcal{N}(B,L) = \int_{\mathbb{S}^{d-1}} &dw \,\,\mathbb{E}_y \bigg[ p_{\nabla L(w)|y}(0) \ \cdot
        \\  & \quad
        \mathbb{E}\left[|\det\nabla^2L(w)| | \nabla L(w)=0,y \right] \bigg] 
    \end{split}
\end{equation} 

To proceed we will need the first couple of derivatives of $L$ (as $w$ varies on $\mathbb{S}^{d-1}$). 
Starting from $\nabla L(w)$, we have 
\begin{align}
\nabla L(w) = \frac{1}{m}\sum_{\mu=1}^{m}\sigma'(w^\intercal Rz_\mu) P_w^\bot Rz_\mu
\end{align}
Noting that the special direction of $w^\intercal R$ appears in all terms  $w^\intercal Rz_\mu$ above, we will apply $P^\bot_w=I-ww^\intercal$, the projection orthogonal to $w$, to express $RR^\intercal$ in a basis, where $w$ is a distinct direction. Thus we have:
\begin{align}
    RR^\intercal \equiv
    \begin{pmatrix}
        w^\intercal RR^\intercal w & w^\intercal RR^\intercal P_w^\bot \\
        P_w^\bot RR^\intercal w    & P_w^\bot RR^\intercal P_w^\bot
    \end{pmatrix}
\end{align}
or, after the introduction of the $w$-dependent parameters $\sqrt{\rho}=\|R^\intercal w\|$, $u=\frac{1}{\sqrt{\rho}}P_w^\bot RR^\intercal w$ and $\tilde{R}= (P_w^\bot RR^\intercal P_w^\bot - uu^\intercal)^{1/2}$:
\begin{align}
    RR^\intercal \equiv
    \begin{pmatrix}
        \rho & \sqrt{\rho} u^\intercal \\
        \sqrt{\rho} u    & \tilde{R}\tilde{R}^\intercal + uu^\intercal
    \end{pmatrix}
\end{align}
Let $Z\in\mathbb{R}^{d\times m}$ be the matrix such that $Z_{i\mu} = (z_\mu)_i$.
Based on the above representation, $w^\intercal RZ \sim \sqrt{\rho} y$ and $P_w^\bot RZ \sim \tilde{R}\Xi +uy^\intercal$, where $y\sim N(0,I_m)$ and $\Xi\sim N(0,I_{(d-1)\times m})$ are independent.
The distribution of $\nabla L$ under this representation becomes
\begin{align}
\label{eq:nablaE_law}
\nabla L(w) \sim \frac{1}{m}\tilde{R}\Xi\sigma'(\sqrt{\rho} y) + \frac{1}{m}y^\intercal\sigma'(\sqrt{\rho} y)u
\end{align}
where we have made use of a notation that extends functions of one variable to functions of vectors: $\sigma(y)_i = \sigma(y_i)$.

Introducing the diagonal matrix $\Lambda:=\Lambda(\sigma''(\sqrt{\rho} y))$ with elements $\Lambda_{\mu,\mu'} := \delta_{\mu,\mu'} \sigma''(\sqrt{\rho} y_\mu)$, we can now express the Hessian of $L(w)$ as
\begin{equation}
    \begin{split}
        \nabla^2 L(w) &\sim \frac{1}{m} (\tilde{R}\Xi + uy^\intercal) \Lambda (\tilde{R}\Xi + uy^\intercal)^\intercal 
        \\ & \qquad\qquad\qquad\qquad
        - \frac{1}{m}\sqrt{\rho} y^\intercal\sigma'(\sqrt{\rho} y) I 
    \end{split}
\end{equation}
where $\Xi$ and $y$ are the same random variables as in \eqref{eq:nablaE_law}

Furthermore, we need the distribution of $\nabla^2 L(w)$ conditioned on both the value of $y$ and the constraint $\nabla L(w) = 0$, i.e. we need to obtain the distribution of $\Xi$ conditioned on $\Xi\sigma'(\sqrt{\rho} y)=\tilde{R}^{-1}uy^\intercal \sigma'(\sqrt{\rho} y)$.
Conditionally on $y$, both $\nabla L(w)$ and $\Xi$ are Gaussian processes and $\nabla L$ is independent from the Gaussian process $\tilde{R}\Xi P_{\sigma'(\sqrt{\rho} y)}^\bot$, 
where $P_{\sigma'(\sqrt{\rho} y)}^\bot$ is the orthogonal projection with respect to the vector $\sigma'(\sqrt{\rho} y)$.
Writing
\begin{equation}
    \begin{split}
        \tilde{R} \Xi &= 
        \tilde{R} \Xi P_{\sigma'(\sqrt{\rho} y)}^\bot + \tilde{R}\Xi \frac{1}{\|\sigma'(\sqrt{\rho} y)\|^2} \sigma'(\sqrt{\rho} y) \sigma'(\sqrt{\rho} y)^\intercal
        \\
        &= \tilde{R} \Xi P_{\sigma'(\sqrt{\rho} y)}^\bot + \frac{m}{\|\sigma'(\sqrt{\rho} y)\|^2}\nabla L(w) \sigma'(\sqrt{\rho} y)^\intercal 
        \\
        & \qquad \qquad \qquad \qquad \qquad 
        -\frac{y^\intercal \sigma'(\sqrt{\rho} y)}{\|\sigma'(\sqrt{\rho} y)\|^2} u\sigma'(\sqrt{\rho} y)^
        \intercal
    \end{split}
\end{equation}
the effect of conditioning becomes obvious.

We thus see that $\nabla^2 L(w)$, conditioned on $\nabla L(w) = 0$ and $y$, is distributed as
\begin{equation}\label{eq:nabla2L}
    \nabla^2 L(w)\Bigg|_{\{\nabla L(w) = 0\},y} 
    \sim \frac{1}{m} (R\Xi + \Pi) \Lambda (R\Xi + \Pi)^\intercal -  \kappa I
\end{equation}
where $\Pi$ is a matrix of rank at most 2 and 
$\kappa = \frac{1}{m}\sqrt{\rho} y^\intercal\sigma'(\sqrt{\rho} y)$.

\subsection{The Expectation of the Determinant} \label{sec:perceptron/rmt}

To calculate $\mathbb{E}\left[|\det\nabla^2L(w)| | \nabla L(w)=0,y \right]$ we will make use of the fact that typically the logarithm of the determinant of a random matrix of the form $H=\frac{1}{m} (R\Xi + \Pi) \Lambda (R\Xi + \Pi)^\intercal$ is dominated by its mean value to an exponential scale:
\begin{equation}
\label{eq:det_assumption_1}
    \log \mathbb{E}_\Xi \left[ |\det(H-\kappa I)| \right] = 
    \mathbb{E}_\Xi\left[\log|\det(H-\kappa I)| \right] + \mathcal{O}(1)
\end{equation}
and that the eigenvalues of $H$ rarely get close enough to zero to obstruct the evaluation of $\mathbb{E}\left[\log|\det(H-\kappa I)|\right]$ by treating it as a spectral statistic through a random matrix limit theorem:
\begin{equation}
\label{eq:det_assumption_2}
    \frac{1}{d}\mathbb{E}_\Xi\left[\log|\det(H-\kappa I)|\right] =
    \mathbb{E}_\lambda\log|\lambda-\kappa| + \mathcal{O}\left(\frac{1}{d}\right)
\end{equation}
where $\lambda$ is distributed according to the (random) limiting spectral measure of $H|_y$.
Probabilistic control of the error terms in \eqref{eq:det_assumption_1} and \eqref{eq:det_assumption_2} to superexponential rate is expected, but remains highly non-trivial from a technical perspective. 
Such control is a highly motivated and common assumption, necessary to proceed with the study of these models \cite{Maillard_Arous_Biroli_2023}.
Attempts at rigorous proofs have been fruitful, yet still unable to tackle a model like ours \cite{Arous_Bourgade_McKenna_2022}.
Nevertheless, the ``mutual information'' $\text{tr}\log(H-\kappa I)$ is an extensively studied object with a well-known Gaussian limiting law with finite variance \cite{Moustakas2003MIMO1, Hachem2006_GaussianCapacityKroneckerProduct, moustakas2013sinr}.

Given the above conjecture, we now only need the spectral measure of $H$ for fixed $y$. 
Clearly the matrix $\Pi$ does not affect the logarithmic potential to first order, and so we will ignore it. 
We will work with the potential defined on $\mathbb{C}^+$ by $G(z) = \mathbb{E}\log(z-\lambda)$, which is connected to the logarithmic potential through $\Re \left[G(\kappa+i\epsilon)\right]\xrightarrow{\epsilon\to0^+}\mathbb{E}\log|\lambda-\kappa|$. 
We know from \cite{paul2009no}, that the potential $G(z)$ can be understood through the Random Matrix Free Energy $\tilde{F}^{RM}$, whose dependence on $\rho$ and $\nu$ will be kept implicit as follows
\begin{equation}
    \begin{split}
        \tilde{F}^{RM}(z;\phi,\bar{\phi}) =&
        \beta\mathbb{E}_\zeta\log\left(1+\frac{1}{\beta} \sigma''(\sqrt{\rho}\zeta)\phi\right)
        \\ & \qquad \qquad 
        +\mathbb{E}_{\lambda_\Sigma}\log\left(z+z\lambda_\Sigma \bar{\phi}\right) +z\phi\bar{\phi}
    \end{split}
\end{equation}
where $\zeta$ is distributed according to the random measure $\nu_y = \frac{1}{m}\sum_\mu \delta_{y_\mu}$ and, as mentioned in Section \ref{sec:results}, $\lambda_\Sigma$ is distributed according to the limiting spectral measure of $\Sigma$.
Specifically we have
\begin{align}
    G(z) = \text{extr}_{\phi,\bar{\phi}}\tilde{F}^{RM}(z;\phi,\bar{\phi})
\end{align}
and the extremal point of $\tilde{F}^{RM}$ is the unique solution of the fixed point equations
\begin{align}\label{eq:phi_bar_eq}
    -z\bar{\phi} &= \mathbb{E} \frac{\sigma''(\sqrt{\rho} \zeta)}{1+\frac{1}{\beta}\sigma''(\sqrt{\rho} \zeta)\phi}
    \\
    -z\phi &= \mathbb{E}\frac{\lambda_\Sigma}{1+\lambda_\Sigma \bar{\phi}} 
\end{align}
In the interest of obtaining a closed form for $\mathbb{E}\log|\lambda-\kappa|$, we can expand the real part of $\tilde{F}^{RM}$ for small values of the imaginary part of the argument of $G(z)$, namely $kappa+i\epsilon$, thus obtaining
\begin{equation}
\label{eq:rmt_perturbed_1}
    \begin{split}
        &\Re\left[\tilde{F}^{RM}(\kappa+i\epsilon;\phi,\bar{\phi})\right] 
        = \beta\mathbb{E}_\zeta\log\left|1+\frac{1}{\beta} \sigma''(\sqrt{\rho} \zeta)\phi\right|
        \\ & \qquad \qquad \qquad 
        + \mathbb{E}_{\lambda_\Sigma}\log\left|\left(\kappa + \kappa \bar{\phi}\lambda_\Sigma\right) \right|
        +\Re[(\kappa+i\epsilon)\bar{\phi}\phi]
    \end{split}
\end{equation}

\subsection{Integration Around the Sphere} \label{sec:perceptron/quadratics}

Next we perform the integration with respect to $w$. 
Starting with the expression of $ p_{\nabla L(w)|y}(0)$, and introducing the order parameter $f=\frac{1}{m}\|\sigma(\sqrt{\rho} y)\|^2$, we have 
\begin{equation}
    p_{\nabla L(w)|y}(0) = (2\pi)^{-d/2}\text{det}\left( \frac{1}{m} f \Sigma\right)^{-1/2} e^{-\frac{\kappa^2}{2\rho}  u^\intercal \left( \frac{1}{m} f \Sigma\right)^{-1} u}
\end{equation}
Recalling that $u=\frac{1}{\sqrt{\rho}}P_w^\bot \Sigma w$, the above can be written as 
\begin{equation}
    \begin{split}
        &\frac{1}{d}\log p_{\nabla L(w)|y}(0) = \frac{1}{2}\log(m) - \frac{1}{2}\log(2\pi f) 
        \\ & \qquad \qquad 
        - \frac{1}{2d}\log\det\left(\Sigma\right)
        -\frac{\beta\kappa^2}{2\rho^2} 
        w^\intercal \Sigma P_w^\bot \left( f \Sigma \right)^{-1} P_w^\bot \Sigma w
    \end{split}
\end{equation}
thus getting
\begin{equation} 
    \begin{split} 
        &\frac{1}{d}\log p_{\nabla L(w)|y}(0) = \frac{1}{2}\log(m) - \frac{1}{2}\log(2\pi f) 
        \\ & \qquad \qquad
        -\frac{1}{2d}\log\text{det}\left(\Sigma\right) 
        +\frac{1}{2}\beta\frac{\kappa^2}{f\rho}\left(1 - \rho  w^\intercal \Sigma^{-1} w \right)
    \end{split} 
\end{equation} 
Since we need to integrate the $w$ dependence over the sphere, we need the joint LDP of $\rho$ and $q=w^\intercal \Sigma^{-1} w$, when $w$ is seen as a uniform random variable on the sphere. The Gaussian analog of this problem is well-studied \cite{bercu2000sharp} and in the spherical case it is easy to see that the rate function $I(\rho,q)$ is given by
\begin{equation} 
    \begin{split} 
    I(\rho,q) &= \sup_{\bar{\rho}\bar{q}}\frac{1}{2d}\log\det\left[
    (1-\bar{\rho}\rho -\bar{q}q)I
    +\bar{\rho}\Sigma
    +\bar{q}\Sigma^{-1}
    \right]
    \\
    &= \sup_{\bar{\rho}\bar{q}}\frac{1}{2}\mathbb{E}_{\lambda_\Sigma}\log\left(
    1-\bar{\rho}\rho -\bar{q}q
    +\bar{\rho}\lambda_\Sigma
    +\bar{q}\frac{1}{\lambda_\Sigma}
    \right)
    \end{split} 
\end{equation} 
where $\lambda_\Sigma$ is a random variable distributed according to the spectral distribution of $\Sigma$. Notably, the function inside the supremum is jointly concave with respect to $\bar{\rho}$ and $\bar{q}$. 

Thus, integrating, with respect to $w$, functions of $\rho$ and $q$ multiplied by $ p_{\nabla L(w)|y}(0)$ can be performed through the LDP with convex rate function
\begin{equation} 
    \begin{split} 
        F^{SP}(\rho,q;\kappa,f) =& \frac{1}{2}\log(\beta e) -\frac{1}{2}\log(f) -\frac{1}{2d}\log\text{det}\left(\Sigma\right)
        \\ & \qquad \qquad \quad
        +\frac{1}{2}\beta\frac{\kappa^2}{f}\left(\frac{1}{\rho} - q \right) - I(\rho,q)
    \end{split} 
\end{equation} 
where the constant term comes from the volume of the $d$-dimensional sphere. 

\subsection{Large Deviations of Empirical Measures} 
\label{sec:perceptron/Sanov}

Finally, we are left with the expectation over $y$, which appears only as statistics of $\zeta$ in the order parameters $\ell$, $\kappa$ and $f$ and in $\tilde{F}^{RM}$. 
Since $\zeta$ is distributed according to the empirical measure $\nu_y$, we can use Sanov's LDP for empirical measures. 
According to Sanov's LDP, the rate function for $\nu_y$ is $-\beta\mathcal{H}(\nu_y|N)$, where $N$ is the one-dimensional normal distribution (since $y$ when unconstrained is Gaussian) and $\mathcal{H}$ is the relative entropy \cite{dembo2009large}. 
To find the infimum over the measure $\nu_y$, we need to take into account that the values of $f$, $\kappa$ and $\ell$ are fixed by the conditions $f=\mathbb{E}_\zeta\sigma'(\zeta)^2$, $\kappa=\mathbb{E}_\zeta\zeta\sigma'(\zeta)$ and $\ell=\mathbb{E}_\zeta\sigma(\zeta)$. 
In addition, $\tilde{F}^{RM}$ includes the expectation $\mathbb{E}_\zeta\log\left|1+\frac{1}{\beta} \sigma''(\sqrt{\rho} \zeta)\phi\right|$. 
Hence, to constrain the measure to fixed values of $\ell$, $\kappa$ and $f$  we introduce auxiliary parameters $\bar{\ell}$, $\bar{\kappa}$ and $\bar{f}$ as Langrange multipliers, and then minimize $\nu_y$ for these parameters fixed, using Sanov's theorem. Subsequently, we can obtain the infimum with respect to these parameters. 
However, care should be taken, since the values of $\phi$, $\bar{\phi}$ depend on these parameters themselves through \eqref{eq:phi_bar_eq}. Thus, starting from \eqref{eq:det_assumption_1} and \eqref{eq:det_assumption_2}, using the representation of \eqref{eq:rmt_perturbed_1} in conjunction with Sanov's LDP,
the quantity inside the log in the left-hand-side in \eqref{eq:det_assumption_1} can be expressed, to first order in $d$, as
\begin{equation}
\label{eq:apply_Sanov}
    \begin{split}
        \log \mathbb{E} &\left[ e^{\text{extr}_{\phi\bar{\phi}}\Re\left[\tilde{F}^{RM}(\kappa+i\epsilon;\phi,\bar{\phi})\right]} \Bigg| f,\kappa,\ell \right] =
        \\ &
        \beta\inf_{\nu} \left(
        \text{extr}_{\phi\bar{\phi}} F^{RM}(\phi,\bar{\phi};\rho,\kappa,\nu,\epsilon)
        + F^{S}(\ell,\kappa,f,\nu;\rho) \right)
    \end{split}
\end{equation} 
where $F^{RM}$ is as defined in \eqref{eq:F_RM} and the infimum is taken over all probability measures $\nu$ parametrized by three real, unconstrained auxiliary parameters, $\bar{\kappa}$, $\bar{f}$ and $\bar{\lambda}$, which fix the values of $f$, $\kappa$ and $\ell$, in the following way:
\begin{equation}
    \nu(d\zeta) 
    = \frac{1}{Z(\bar{\ell},\bar{\kappa},\bar{f};\rho)} e^{-\frac{1}{2}\zeta^2-\frac{1}{\beta}F_\nu(\zeta;\bar{\ell},\bar{\kappa},\bar{f},\rho)} d\zeta
\end{equation}
$Z$ is defined above through the normalization of $\nu\zeta)$, while $F_\nu$ is given by \eqref{eq:data_bias}, which we replicate below for concreteness:
\begin{equation}
    \begin{split} 
    F_\nu(\zeta;\bar{\ell},\bar{\kappa},\bar{f},\rho) = \bar{\ell}\sigma(\sqrt{\rho}\zeta)
    &+\bar{\kappa}\sqrt{\rho} \zeta\sigma'(\sqrt{\rho}\zeta)
    \\ &+\bar{f}\sigma'(\sqrt{\rho}\zeta)^2
    \end{split} 
\end{equation}
and the Sanov LDP gives the total rate function $F^{S}$ as defined in \eqref{eq:sanov_and_lagrange}
\begin{equation}
    F^{S}(\nu;\ell,\kappa,f,\rho)  = \bar{\ell}\ell +\bar{\kappa}\kappa +\bar{f}f +  \beta\log Z(\bar{\ell},\bar{\kappa},\bar{f};\rho)
\end{equation}

The change from $\tilde{F}^{RM}$ to $F^{RM}$ presented in \eqref{eq:apply_Sanov} results from taking the infimum over $\nu$ within the extremization over $\phi$ and $\bar{\phi}$, in order to calculate the expected value $\mathbb{E}\log\left|1+\frac{1}{\beta} \sigma''(\sqrt{\rho} \zeta)\phi\right|$ in \eqref{eq:rmt_perturbed_1}.
This amounts to replacing that expectation with $\log\mathbb{E}\left|1+\frac{1}{\beta} \sigma''(\sqrt{\rho} \zeta)\phi\right|$ and adding the rate function $F^{S}$ to the total LDP. 

Taking this and the previous subsections into account we arrive at \eqref{eq:N} and its discussion as presented in Section \ref{sec:results}. 

\section{Generalized Linear Models: $L_1(w)$} 
\label{sec:glm}

In this section we will explain the main changes in the above procedure when calculating $\mathbb{E}\mathcal{N}(U,B;L_1)$ instead of $\mathbb{E}\mathcal{N}(U,B;L)$. 
In this case, in addition to $w^\intercal R z_\mu$, a second set of Gaussian random variables appears, namely $w_*^\intercal R z_\mu$.
The correlations between these two sets of random variables is non-zero only on the same data index
\begin{equation}
    \mathbb{E}\left[ \left( w^\intercal R z_\mu \right) \left( w_*^\intercal R z_{\mu'} \right) \right] = \delta_{\mu\mu'} w_*^\intercal RR^\intercal w 
\end{equation}
Now the distribution of the loss can be written in terms of two independent normal $m$-dimensional vectors $y$ and $y'$ in the following form
\begin{align}
    L_1(w) \sim \frac{1}{m} \|\sigma(\sqrt{\rho} y) - \sigma(\sqrt{p}y+\sqrt{\rho_*-p}y')\|^2
\end{align}
We thus see the appearance of a new order parameter, $\sqrt{p}=w_*^\intercal RR^\intercal w/\sqrt{\rho}$, as well as the (fixed) variable $\sqrt{\rho_*}=\|R^\intercal w_*\|$. Furthermore the gradient over $L_1(w)$ can be expressed as,
\begin{equation}
    \begin{split}
        \nabla L_1(w) \sim \frac{1}{m}\|\theta_\sigma(\sqrt{\rho} y,\sqrt{p} y + &\sqrt{\rho_*-p}y')\|\tilde{R}\xi 
        \\ & \qquad
        +\frac{\kappa}{\sqrt{\rho}} u
        +\frac{\kappa'}{\sqrt{\rho}} v
    \end{split}
\end{equation}
with $\tilde{R}= (P_w^\bot RR^\intercal P_w^\bot - uu^\intercal - vv^\intercal)^{1/2}$, $u=\frac{1}{\sqrt{\rho}}P_w^\bot RR^\intercal w$ and $v=\frac{1}{\sqrt{\rho_*-p}}P_w^\bot RR^\intercal\left(w^*-\frac{\sqrt{p}}{\sqrt{\rho}}w\right)$, where for concreteness we defined the following function
\begin{equation}
    \theta_\sigma(z,z')   = (\sigma(z) - \sigma(z'))\sigma'(z)
\end{equation}
and now we have two $\kappa$-varaibles, namely
\begin{align}
    \begin{split}
    \kappa &= \frac{1}{m}\sum_{\mu=1}^{m}
    (\sigma(\sqrt{\rho} y_\mu) - \sigma(\sqrt{p} y_\mu + \sqrt{\rho_*-p}y'_\mu))\cdot
    \\
    & \qquad\qquad\qquad\qquad\qquad\qquad\quad\sigma'(\sqrt{\rho} y_\mu) \sqrt{\rho} y_\mu
    \end{split}
    \\
    \begin{split}
    \kappa' &= \frac{1}{m}\sum_{\mu=1}^{m}(\sigma(\sqrt{\rho} y_\mu) - \sigma(\sqrt{p} y_\mu + \sqrt{\rho_*-p}y'_\mu))\cdot
    \\
    &\qquad\qquad\qquad\qquad\qquad\qquad\quad\sigma'(\sqrt{\rho} y_\mu) \sqrt{\rho} y'_\mu
    \end{split}
\end{align}
Finally, for the Hessian we have, up to low rank perturbations,
\begin{align}
    [\nabla^2 L_1(w)|\nabla L_1(w)=0] &\sim \frac{1}{m} R\Xi \Lambda \Xi^{\intercal}R^{\intercal} - \kappa I
\end{align}
where  
\begin{equation}
    \Lambda_{\mu,\mu'} := \delta_{\mu,\mu'} \psi_\sigma(\sqrt{\rho} y_\mu,\sqrt{p} y_\mu + \sqrt{\rho_*-p}y'_\mu),
\end{equation} 
using the definition of 
\begin{equation}
    \psi_\sigma(z,z')    = (\sigma(z) - \sigma(z'))\sigma''(z) 
\end{equation}

Under this change of $\Lambda$, the same Random Matrix Theory result as above is applicable. The real part of the corresponding free energy takes, up to first order in $\epsilon$, the following form
\begin{equation}
\label{eq:tildeF1_RM}
    \begin{split}
        &\Re[\tilde{F}^{RM}_1(\rho,\kappa+i\epsilon;\phi,\bar{\phi})]
        = \mathbb{E}_{\lambda_\Sigma}\log\left|\left(\kappa + \kappa \bar{\phi}\lambda_\Sigma\right) \right|
        \\ & \qquad  
        +\beta\mathbb{E}_{\zeta,\zeta'}\log\left|1+\frac{1}{\beta} \psi_\sigma(\sqrt{\rho}\zeta,\sqrt{p}\zeta + \sqrt{\rho_*-p}\zeta')\phi\right|
        \\ & \qquad \qquad \qquad \qquad \qquad \qquad 
        +\Re((\kappa+i\epsilon)\bar{\phi} \phi) 
    \end{split}        
\end{equation}
where now the random variable $\zeta'$ also appears, which is distributed according to the empirical distribution of the coordinates of $y'$.

Furthermore, the form of $\nabla L_1(w)$ changes the probability density at 0 to
\begin{equation}
    \begin{split}
        &\frac{1}{d}\log p_{\nabla L(w)|y}(0) = \frac{1}{2}\log(m) - \frac{1}{2}\log(2\pi f)  
        \\ & \qquad \qquad \qquad \qquad \qquad \qquad \qquad 
        - \frac{1}{2d}\log\text{det}\left(\Sigma\right) 
        \\ & \quad
        +\frac{1}{2}\beta\frac{1}{f\rho}\left( (1-q\rho)\kappa^2 +\kappa'^2 - 2\frac{(w_*^\intercal w) \sqrt{\rho}-\sqrt{p}}{\sqrt{\rho_*-p}}\kappa\kappa' \right)
    \end{split}
\end{equation}
so now we need a joint LDP for $\rho$, $q$, $p$ as well as $r=w_*^\intercal w$, which is characterized by the rate function $I_1(\rho,q,p,r)$ given by
\begin{equation}
    \begin{split}
        &I_1(\rho,q,p,r) 
        \\ & \qquad 
        = \sup_{\bar{\rho},\bar{q},\bar{p},\bar{r}} \frac{1}{2d}\text{tr}\log\left[
        (1-\bar{\rho}\rho -\bar{q}q-\bar{r}r^2-\bar{p}p)I \right.
        \\ & \qquad \qquad \qquad  
        \left. 
        +\bar{\rho}\Sigma
        +\bar{q}\Sigma^{-1} - \bar{r}w_*w_*^\intercal -\frac{\bar{p}}{\rho}\left(\Sigma w_*w_*^\intercal \Sigma\right)
        \right]
    \end{split}
\end{equation}
As a result, the geometric part of the free energy takes the form
\begin{equation}
\label{eq:F1_SP}
    \begin{split}
        &F^{SP}_1(\rho,q,r,p;\kappa,\kappa',f) = \frac{1}{2}\log(\beta e)
        - \frac{1}{2d}\log\text{det}\left(\Sigma\right) 
        \\ & \quad
        +\frac{1}{2}\beta\frac{1}{f\rho}\left( (1-q\rho)\kappa^2 +\kappa'^2 - 2\frac{(w_*^\intercal w) \sqrt{\rho}-\sqrt{p}}{\sqrt{\rho_*-p}}\kappa\kappa' \right)
        \\ & \qquad \qquad \qquad \qquad \qquad \qquad \qquad \qquad 
        -I_1(\rho,q,p,r) 
    \end{split}
\end{equation}

Finally, to complete the large deviation analysis one has to take into account the joint deviation of the densities of $\zeta$ and $\zeta'$, and specifically with respect to the loss, $\ell$, as well as $\kappa$, $\kappa'$, $f$ and the expected value appearing in \eqref{eq:tildeF1_RM}.
Utilizing Sanov's theorem as we did in Section \ref{sec:perceptron/Sanov}, the Random Matrix free energy finally takes the form
\begin{equation}
\label{eq:F1_RM}
    \begin{split}
        &F^{RM}_1(\phi,\bar{\phi};\rho,p,\kappa,\nu,\epsilon)]
        = \mathbb{E}_{\lambda_\Sigma}\log\left|\left(\kappa + \kappa \bar{\phi}\lambda_\Sigma\right) \right|
        \\ & \qquad 
        +\beta\log\mathbb{E}_{\zeta,\zeta'}\left|1+\frac{1}{\beta} \psi_\sigma(\sqrt{\rho}\zeta,\sqrt{p}\zeta + \sqrt{\rho_*-p}\zeta')\phi\right|
        \\ & \qquad \qquad \qquad \qquad \qquad \qquad \qquad 
        +\Re((\kappa+i\epsilon)\bar{\phi} \phi)
    \end{split}
\end{equation}
with $\zeta$ and $\zeta'$ being distributed according to the distribution $\nu$ with density
\begin{equation}
    \begin{split}
        &\nu(d\zeta d\zeta') = \frac{1}{Z_1(\bar{\ell},\bar{\kappa},\bar{\kappa}',\bar{f};\rho,p)}
        \exp\left\{-\frac{1}{2}\zeta^2-\frac{1}{2}\zeta'^2
        \right.
        \\ & \qquad \qquad \qquad
        - \frac{1}{\beta}F_1^\nu(\zeta,\zeta';\bar{\ell},\bar{\kappa},\bar{\kappa}',\bar{f},\rho,p)
        \Bigg\} d\zeta d\zeta'
    \end{split}
\end{equation}
where $Z_1$ is determined by normalization and
\begin{equation}
    \begin{split}
        &F_1^\nu(\zeta,\zeta';\bar{\ell},\bar{\kappa},\bar{\kappa}',\bar{f},p,\rho) =
        \\ & \qquad \qquad \qquad \quad 
        \bar{\ell}(\sigma(\sqrt{\rho} \zeta)-\sigma(\sqrt{p}\zeta+\sqrt{\rho_*-p}\zeta'))^2
        \\ & \qquad \qquad \qquad 
        +\bar{f}(\theta_\sigma(\sqrt{\rho} \zeta,\sqrt{p} \zeta + \sqrt{\rho_*-p}\zeta')^2
        \\ & 
        +\bar{\kappa}\left((\sigma(\sqrt{\rho} \zeta) - \sigma(\sqrt{p} \zeta + \sqrt{\rho_*-p}\zeta'))\sigma'(\sqrt{\rho} \zeta) \sqrt{\rho} \zeta)\right)
        \\ & 
        +\bar{\kappa}\left((\sigma(\sqrt{\rho} \zeta) - \sigma(\sqrt{p} \zeta + \sqrt{\rho_*-p}\zeta'))\sigma'(\sqrt{\rho} \zeta) \sqrt{\rho} \zeta')\right)
    \end{split}
\end{equation}
Concluding, the barred parameters are to be determined by the following rate function 
\begin{equation}
\label{F1_Sh}
    \begin{split}
        F^{S}_1(\ell,\kappa,\kappa',f,\nu;\rho,p) = &\bar{f}f +\bar{\kappa}\kappa +\bar{\kappa}'\kappa' +\bar{\ell}\ell 
        \\ & \qquad 
        +\beta\log Z_1(\bar{\ell},\bar{\kappa},\bar{\kappa}',\bar{f},p,\rho)
    \end{split}
\end{equation}

\section{Discriminating Perceptron: $L_2(w)$} 
\label{sec:discriminator} 

We will now briefly present the significant steps of the calculation of $\mathbb{E}\mathcal{N}(U,B;L_2)$.
Here the distribution of $L_2$ is of the form
\begin{align}
    L_2(w) \sim \frac{1}{m_1}\sum_{\mu=1}^{m_1}\sigma_1(\sqrt{\rho}_1y_{1\mu}) 
    -\frac{1}{m_2}\sum_{\mu=1}^{m_2}\sigma_2(\sqrt{\rho}_2y_{2\mu})
\end{align} 
where now for $j=1,2$, $y_j$ is an $m_j$-dimensional normally distributed vector and $\sqrt{\rho_j}:=\|R_j^\intercal w\|$.
It should be pointed out that in this section, all variables that were defined and appearing in the analysis of Section \ref{sec:perceptron} (with the exception of $\epsilon$) appear in pairs, such as $\rho_1$ and $\rho_2$.  When possible we will refer to them jointly, dropping the index $1,2$, e.g. $\rho = (\rho_1,\rho_2)$.
The distribution of $\nabla L_2(w)$ takes the form
\begin{equation}
    \begin{split}
        \nabla L_2(w) \sim& \frac{1}{m_1}\tilde{R}_1\Xi_1\sigma'(\sqrt{\rho}_1y_1) - \frac{1}{m_2}\tilde{R}_2\Xi_2\sigma'(\sqrt{\rho}_2y_2) 
        \\ & \ 
        + \frac{1}{m_1}y_1^\intercal\sigma'(\sqrt{\rho}_1y_1)u_1 - \frac{1}{m_2}y_2^\intercal\sigma'(\sqrt{\rho}_2y_2)u_2
    \end{split}
\end{equation}
with $u_j=\frac{1}{\sqrt{\rho}}P_w^\bot RR^\intercal w$ and $\tilde{R}_j= (P_w^\bot R_jR_j^\intercal P_w^\bot - u_ju_j^\intercal)^{1/2}$ and $\Xi_j$ a matrix with independent normal entries.

Finally, introducing $\kappa_j = \frac{1}{m_j}\sqrt{\rho_j} y_j^\intercal R_jR_j'(\sqrt{\rho_j} y_j)$ and the matrix $\Lambda_j$ with entries $\Lambda_{j,\mu,\mu'} := \delta_{\mu,\mu'} \sigma''(\sqrt{\rho_j} y_{j\mu})$ the Hessian is distributed, up to low rank perturbations, according to the following random matrix
\begin{equation}
    \begin{split}
        &[\nabla^2 L_2(w)|\nabla L_2(w)=0] \sim \frac{1}{m_1} R_1\Xi_1 \Lambda_1 \Xi_1^\intercal R_1^\intercal
        \\ & \qquad \qquad \qquad \qquad 
        - \frac{1}{m_2} R_2\Xi_2 \Lambda_2 \Xi_2^\intercal R_2^\intercal
        - (\kappa_1-\kappa_2) I
    \end{split}
\end{equation}
This competitive structure in the Hessian necessitates a more general Random Matrix Theory result, which can be found in \cite{Moustakas2007_MIMO1, Couillet_Debbah_Silverstein_2011}. Applying this result, as we did in Section \ref{sec:perceptron/rmt}, we obtain, to first order in $\epsilon$, the following free energy
\begin{equation}
\label{eq:tildeF2_RM}
    \begin{split}
        &\Re\left[\tilde{F}^{RM}(\phi_1,\phi_2,\bar{\phi}_1,\bar{\phi}_2;\rho,\kappa_1-\kappa_2+i\epsilon)\right] = 
        \\ & \qquad \qquad \qquad 
        \frac{\beta}{\alpha_1}\mathbb{E}_{\zeta_1}\log\left|1+\frac{\alpha_1}{\beta} \sigma_1''(\sqrt{\rho}_1 \zeta_1)\phi_1\right|
        \\ & \qquad \qquad \qquad 
        +\frac{\beta}{\alpha_2}\mathbb{E}_{\zeta_2}\log\left|1-\frac{\alpha_2}{\beta} \sigma_2''(\sqrt{\rho}_2 \zeta_2)\phi_2\right|
        \\ & \qquad 
        +\frac{1}{d}\log\bigg|\det\bigg(
        (\kappa_1-\kappa_2)I 
        \\ & \qquad \qquad \qquad \qquad 
        +(\kappa_1-\kappa_2)\bar{\phi}_1G -(\kappa_1-\kappa_2)\bar{\phi}_2\Sigma
        \bigg)\bigg|
        \\ & \qquad \qquad \qquad \quad 
        +\Re((\kappa_1-\kappa_2+i\epsilon)(\bar{\phi}_1 \phi_1 -\bar{\phi}_2 \phi_2))
    \end{split}
\end{equation} 
where we introduced the random variables $\zeta_1$, $\zeta_2$ distributed, respectively, according to $\nu_1(y_1)=\sum_{\mu=1}^{m_1}\delta_{y_{1\mu}}$ and $\nu_2(y_2)=\sum_{\mu=1}^{m_2}\delta_{y_{2\mu}}$.

We now move to the geometric order parameters, introducing $f_j=\frac{1}{m_j}\|\sigma'(\sqrt{\rho_j} y_j)\|^2$, and the matrices 
\begin{align}
    S &=\alpha_1 f_1 G + \alpha_2 f_2 \Sigma, 
    \\ 
    D &= \frac{\kappa_1}{\rho_1}G - \frac{\kappa_2}{\rho_2}\Sigma
\end{align}
We see that the probability density of $\nabla L_2(w)$ at $0$ equals
\begin{equation}
    \begin{split}
        &\frac{1}{d}\log p_{\nabla L(w)|y}(0) = \frac{1}{2}\log(m_1+m_2) - \frac{1}{2}\log(2\pi) 
        \\ & 
        - \frac{1}{2d}\log\det S 
        -\frac{\beta}{2} w^\intercal
        D S^{-1} D w
        -\frac{\beta}{2}
        \left(\kappa_1 -\kappa_2\right)^2 w^\intercal S^{-1}w 
        \\ & \qquad \qquad \qquad \qquad \qquad \qquad 
        +\beta \left(\kappa_1 -\kappa_2\right) w^\intercal S^{-1} D w  
    \end{split}
\end{equation}
from which we can see that 
\begin{equation}
    \begin{split}
        &F^{SP}_2(\rho,q;\kappa,f) = \frac{1}{2}\log(\beta e)
        - \frac{1}{2d}\log\det S  -\frac{\beta}{2} q_1
        \\ & \qquad \quad 
        -\frac{\beta}{2} q_2
        \left(\kappa_1 -\kappa_2\right)^2 
        +\beta q_3
        \left(\kappa_1 -\kappa_2\right)
        -I_2(q,\rho;\kappa,f) 
    \end{split}
\end{equation}
In the above, we have made use of the joint rate function $I_2$ for the variables $\rho_j$ and
\begin{align}
    q_1 &= w^\intercal D S^{-1} D w,
    \\
    q_2 &= w^\intercal S^{-1} w, 
    \\
    q_3 &= w^\intercal S^{-1} Dw
\end{align}
The rate function of such an LDP is of the form
\begin{equation}
    \begin{split}
        &I_2(\rho,q;f,\kappa) = \sup_{\bar{\rho_1},\bar{\rho_2},\bar{q_1},\bar{q_2},\bar{q_3}} \Bigg(\frac{1}{2d}\log\det\Bigg[\bar{\rho}_1G +\bar{\rho}_2\Sigma
        \\ & \qquad \quad 
        +(1-\bar{\rho}_1\rho_1 -\bar{\rho}_2\rho_2 -\bar{q}_1q_1 -\bar{q}_2q_2 -\bar{q}_3q_3)I 
        \\ & \qquad \qquad  
        +\bar{q}_1 D S^{-1} D
        +\bar{q}_2 S^{-1}
        +\frac{\bar{q}_3}{2}\left( S^{-1} D +h.c. \right)
        \Bigg]\Bigg)
    \end{split}
\end{equation}

Throughout the calculation until now, the random vectors $y_1$ and $y_2$ determine only averages of functions of $\zeta_1$ and $\zeta_2$ through the pair of $\kappa$ variables, the pair of $f$ variables and the two expectations in \eqref{eq:tildeF2_RM}.
Using Sanov's LDP we can get a rate function for these statistics and the loss, $\ell$, of the form 
\begin{equation}
    \begin{split}
        &F^{S}_2(\ell,\kappa,f,\nu;\rho) = 
        \bar{\ell}\ell
        +\bar{f}_1f_1 +\bar{f}_2f_2 
        +\bar{\kappa}_1\kappa_1 +\bar{\kappa}_2\kappa_2
        \\ & \ \ 
        +\frac{\beta}{\alpha_1}\log Z_{\alpha_1}(\bar{\ell},\bar{f}_1,\bar{\kappa}_1;\rho_1)
        +\frac{\beta}{\alpha_2}\log Z_{\alpha_2}(-\bar{\ell},\bar{f}_2,\bar{\kappa}_2;\rho_2) 
    \end{split}
\end{equation} 
with 
\begin{equation} 
    Z_{\alpha}(\bar{\ell},\bar{f},\bar{\kappa};\rho) 
    = \int d\zeta \exp \Bigg\{ -\frac{1}{2}\zeta^2 
    -\frac{\alpha}{\beta}F_\nu(\zeta;\bar{\ell},\bar{\kappa},\bar{f},\rho)
    \Bigg\}
\end{equation} 
where $F_\nu$ is as in \eqref{eq:data_bias} 

When we apply Sanov's LDP, the random matrix LDP takes the form
\begin{equation}
\label{eq:tildeF2_RM_final}
    \begin{split}
        &F^{RM}(\phi,\bar{\phi};\rho,\kappa,\epsilon) = 
        \frac{\beta}{\alpha_1}\log\mathbb{E}_{\zeta_1}\left|1+\frac{\alpha_1}{\beta} \sigma_1''(\sqrt{\rho}_1 \zeta_1)\phi_1\right|
        \\ & \qquad \qquad \qquad \quad 
        +\frac{\beta}{\alpha_2}\log\mathbb{E}_{\zeta_2}\left|1-\frac{\alpha_2}{\beta} \sigma_2''(\sqrt{\rho}_2 \zeta_2)\phi_2\right|
        \\ & \qquad 
        +\frac{1}{d}\log\bigg|\det\bigg(
        (\kappa_1-\kappa_2)I 
        \\ & \qquad \qquad \qquad \qquad 
        +(\kappa_1-\kappa_2)\bar{\phi}_1G -(\kappa_1-\kappa_2)\bar{\phi}_2\Sigma
        \bigg)\bigg|
        \\ & \qquad \qquad \qquad \qquad 
        +\Re((\kappa_1-\kappa_2+i\epsilon)(\bar{\phi}_1 \phi_1 -\bar{\phi}_2 \phi_2))
    \end{split}
\end{equation}
where $\zeta_1$ has density $\nu_1$ given through
\begin{align}
    \nu_1(d\zeta_1) \propto
    \exp \Bigg\{ -\frac{1}{2}\zeta^2 
    -\frac{\alpha_1}{\beta}F_\nu(\zeta_1;\bar{\ell},\bar{\kappa}_1,\bar{f}_1,\rho_1)
    \Bigg\}
    d\zeta_1
\end{align}
and $\zeta_2$ is distributed similarly, but replacing $\bar{\ell}$ with $-\bar{\ell}$.

\section{Conclusion - Discussion}

In conclusion, we have applied the methodology introduced in \cite{Maillard_Arous_Biroli_2023}  to the case of the landscape complexity of empirical risk of a number of generalized linear models, all in the case where the high-dimensional input data are correlated random Gaussian vectors. The motivation behind correlations is the fact that real-life data have underlying structure, including sparsity. As a simpler problem we started by analyzing in detail the loss function of the form given in \eqref{eq:L}, focusing on the presence of the covariance matrix. Subsequently, we included a preferred direction in the sphere, corresponding to the planted vector, in a teacher-student setting, and finally we introduced a loss model corresponding to a discriminating perceptron, where two sets of (differently) correlated data vectors are fed to the perceptron, which tries to discriminate between them.

To check the validity of our methodology results, we have analytically evaluated the quadratic loss case where $\sigma(x)=\frac{x^2}{2}$. We have found that the complexity is zero inside the bulk density of eigenvalues of the (semicorrelated) Wishart matrix \eqref{eq:nabla2L} (with $\Lambda=I$) and negative outside, showcasing the fact that in this case the critical points on the sphere correspond to the eigenvalues of the above matrix. It is interesting to point out that the quadratic function is the only one for which (any of) the three functionals ($\zeta\sigma'(\zeta)$, $\sigma'(\zeta)^2$ and $\sigma(\zeta)$) that determine the  optimized density $\nu(d\zeta)$, coincide.

In the analysis of the first two models, an important generalization to previous results is the fact that, due the presence of the data correlations, the Hessian is now a doubly correlated Wishart matrix. In contrast, the third discriminating perceptron model has a Hessian with a difference of two doubly correlated Wishart matrices.

An important output of our analysis is that the data correlations introduces geometrical constraints on the sphere. Specifically, in the simpler first case we analyzed corresponding to loss function given by \eqref{eq:L}, we need to obtain a joint LDP for the quantities $\rho=w^T\Sigma w$ and $q=w^T\Sigma^{-1} w$, which creates non-uniformities in the space of $w$. This effect is exacerbated in the other two models we analyzed. In the case of $L_1(w)$, the existence of a preferred direction $w_*$, introduces two additional overlaps that need to be taken into account, namely $p=(w_*^T\Sigma w)^2/\rho$ and of course $r=w^T_*w$. The more interesting situation arises in the case of the discriminatory perceptron, in which, although without a preferred direction, we need to find joint LDPs, not only for the quantities $\rho_1=w^T\Sigma w$ and $\rho_2=w^TG w$, but also for four geometric objects corresponding to ellipsoids, which combine both correlation matrices $\Sigma$ and $G$ in non-trivial ways. Hence, the optimization procedure over such constrained geometries is very likely nonconvex.

Interestingly, the above analysis can provide a link between the empirical and the generalization loss of each model. For simplicity, we discuss the case of the first model with loss function $L(w)$. When the value of $\ell$ is such that the critical points of the model have a Hessian with a finite fraction of negative eigenvalues, a local algorithm such as gradient descent and its variants will easily find directions to descend to lower values of $\ell$. However, at the value of $\ell$, where the bulk of the Hessian spectrum becomes non-negative (notwithstanding a finite number of negative eigenvalues, with eigendirections that are hard to find),  local algorithms are expected to slow down and at worst get stuck. Based on the analysis in this paper, that value of $\ell^*$ will correspond to a specific value of $\rho=w^T\Sigma w$. At the same time, we know that the generalization error $\mathbb{E}_z\sigma(w^T\Sigma^{1/2}z)$ can be expressed uniquely in terms as $\mathbb{E}_x\sigma(x\sqrt{\rho})$, where $x$ is a unit variance Gaussian variable. Hence, the generalization error can be estimated uniquely at this value of $\rho(\ell^*)$. Similar considerations can made for the generalization errors of the other models discussed here, which depend on the values of other ``order parameters'' at the value of $\ell$, where the system gets stuck.

Building on our results in this paper, we plan to obtain careful numerical results of the optimization problems discussed in the three model-cases here. This task is not trivial, given the non-convexity that appears in the solution spaces. In addition, the complexity of the region of $w$ which has a Hessian with a finite and zero index can be obtained by carefully analyzing the large deviation of the minimum eigenvalues of the Hessian function, as was done in \cite{Auffinger, Choromanska_LeCun_Arous_2015}. Hopefully then, the domain of attraction of the resulting local minima can also be assessed.


\begin{thebibliography}{56}%
\makeatletter
\providecommand \@ifxundefined [1]{%
 \@ifx{#1\undefined}
}%
\providecommand \@ifnum [1]{%
 \ifnum #1\expandafter \@firstoftwo
 \else \expandafter \@secondoftwo
 \fi
}%
\providecommand \@ifx [1]{%
 \ifx #1\expandafter \@firstoftwo
 \else \expandafter \@secondoftwo
 \fi
}%
\providecommand \natexlab [1]{#1}%
\providecommand \enquote  [1]{``#1''}%
\providecommand \bibnamefont  [1]{#1}%
\providecommand \bibfnamefont [1]{#1}%
\providecommand \citenamefont [1]{#1}%
\providecommand \href@noop [0]{\@secondoftwo}%
\providecommand \href [0]{\begingroup \@sanitize@url \@href}%
\providecommand \@href[1]{\@@startlink{#1}\@@href}%
\providecommand \@@href[1]{\endgroup#1\@@endlink}%
\providecommand \@sanitize@url [0]{\catcode `\\12\catcode `\$12\catcode `\&12\catcode `\#12\catcode `\^12\catcode `\_12\catcode `\%12\relax}%
\providecommand \@@startlink[1]{}%
\providecommand \@@endlink[0]{}%
\providecommand \url  [0]{\begingroup\@sanitize@url \@url }%
\providecommand \@url [1]{\endgroup\@href {#1}{\urlprefix }}%
\providecommand \urlprefix  [0]{URL }%
\providecommand \Eprint [0]{\href }%
\providecommand \doibase [0]{http://dx.doi.org/}%
\providecommand \selectlanguage [0]{\@gobble}%
\providecommand \bibinfo  [0]{\@secondoftwo}%
\providecommand \bibfield  [0]{\@secondoftwo}%
\providecommand \translation [1]{[#1]}%
\providecommand \BibitemOpen [0]{}%
\providecommand \bibitemStop [0]{}%
\providecommand \bibitemNoStop [0]{.\EOS\space}%
\providecommand \EOS [0]{\spacefactor3000\relax}%
\providecommand \BibitemShut  [1]{\csname bibitem#1\endcsname}%
\let\auto@bib@innerbib\@empty
\bibitem [{\citenamefont {LeCun}\ \emph {et~al.}(1998)\citenamefont {LeCun}, \citenamefont {Bottou}, \citenamefont {Bengio},\ and\ \citenamefont {Haffner}}]{lecun1998gradient}%
  \BibitemOpen
  \bibfield  {author} {\bibinfo {author} {\bibfnamefont {Y.}~\bibnamefont {LeCun}}, \bibinfo {author} {\bibfnamefont {L.}~\bibnamefont {Bottou}}, \bibinfo {author} {\bibfnamefont {Y.}~\bibnamefont {Bengio}}, \ and\ \bibinfo {author} {\bibfnamefont {P.}~\bibnamefont {Haffner}},\ }\href@noop {} {\bibfield  {journal} {\bibinfo  {journal} {Proceedings of the IEEE}\ }\textbf {\bibinfo {volume} {86}},\ \bibinfo {pages} {2278} (\bibinfo {year} {1998})}\BibitemShut {NoStop}%
\bibitem [{\citenamefont {Xiao}\ \emph {et~al.}(2017)\citenamefont {Xiao}, \citenamefont {Rasul},\ and\ \citenamefont {Vollgraf}}]{xiao2017fashion}%
  \BibitemOpen
  \bibfield  {author} {\bibinfo {author} {\bibfnamefont {H.}~\bibnamefont {Xiao}}, \bibinfo {author} {\bibfnamefont {K.}~\bibnamefont {Rasul}}, \ and\ \bibinfo {author} {\bibfnamefont {R.}~\bibnamefont {Vollgraf}},\ }\href@noop {} {\bibfield  {journal} {\bibinfo  {journal} {arXiv preprint arXiv:1708.07747}\ } (\bibinfo {year} {2017})}\BibitemShut {NoStop}%
\bibitem [{\citenamefont {Li}\ \emph {et~al.}(2018)\citenamefont {Li}, \citenamefont {Xu}, \citenamefont {Taylor}, \citenamefont {Studer},\ and\ \citenamefont {Goldstein}}]{li2018visualizing}%
  \BibitemOpen
  \bibfield  {author} {\bibinfo {author} {\bibfnamefont {H.}~\bibnamefont {Li}}, \bibinfo {author} {\bibfnamefont {Z.}~\bibnamefont {Xu}}, \bibinfo {author} {\bibfnamefont {G.}~\bibnamefont {Taylor}}, \bibinfo {author} {\bibfnamefont {C.}~\bibnamefont {Studer}}, \ and\ \bibinfo {author} {\bibfnamefont {T.}~\bibnamefont {Goldstein}},\ }\href@noop {} {\bibfield  {journal} {\bibinfo  {journal} {Advances in neural information processing systems}\ }\textbf {\bibinfo {volume} {31}} (\bibinfo {year} {2018})}\BibitemShut {NoStop}%
\bibitem [{\citenamefont {Choromanska}\ \emph {et~al.}(2015{\natexlab{a}})\citenamefont {Choromanska}, \citenamefont {Henaff}, \citenamefont {Mathieu}, \citenamefont {Arous},\ and\ \citenamefont {LeCun}}]{Choromanska_Henaff_Mathieu_Arous_LeCun_2015}%
  \BibitemOpen
  \bibfield  {author} {\bibinfo {author} {\bibfnamefont {A.}~\bibnamefont {Choromanska}}, \bibinfo {author} {\bibfnamefont {M.}~\bibnamefont {Henaff}}, \bibinfo {author} {\bibfnamefont {M.}~\bibnamefont {Mathieu}}, \bibinfo {author} {\bibfnamefont {G.~B.}\ \bibnamefont {Arous}}, \ and\ \bibinfo {author} {\bibfnamefont {Y.}~\bibnamefont {LeCun}},\ }in\ \href {https://proceedings.mlr.press/v38/choromanska15.html} {\emph {\bibinfo {booktitle} {Proceedings of the Eighteenth International Conference on Artificial Intelligence and Statistics}}}\ (\bibinfo  {publisher} {PMLR},\ \bibinfo {year} {2015})\ p.\ \bibinfo {pages} {192–204}\BibitemShut {NoStop}%
\bibitem [{\citenamefont {Baskerville}\ \emph {et~al.}(2022)\citenamefont {Baskerville}, \citenamefont {Keating}, \citenamefont {Mezzadri},\ and\ \citenamefont {Najnudel}}]{Baskerville_Keating_Mezzadri_Najnudel_2022}%
  \BibitemOpen
  \bibfield  {author} {\bibinfo {author} {\bibfnamefont {N.~P.}\ \bibnamefont {Baskerville}}, \bibinfo {author} {\bibfnamefont {J.~P.}\ \bibnamefont {Keating}}, \bibinfo {author} {\bibfnamefont {F.}~\bibnamefont {Mezzadri}}, \ and\ \bibinfo {author} {\bibfnamefont {J.}~\bibnamefont {Najnudel}},\ }\href {\doibase 10.1007/s10955-022-02875-w} {\bibfield  {journal} {\bibinfo  {journal} {Journal of Statistical Physics}\ }\textbf {\bibinfo {volume} {186}},\ \bibinfo {pages} {29} (\bibinfo {year} {2022})}\BibitemShut {NoStop}%
\bibitem [{\citenamefont {Maillard}\ \emph {et~al.}(2020)\citenamefont {Maillard}, \citenamefont {Arous},\ and\ \citenamefont {Biroli}}]{Maillard_Arous_Biroli_2023}%
  \BibitemOpen
  \bibfield  {author} {\bibinfo {author} {\bibfnamefont {A.}~\bibnamefont {Maillard}}, \bibinfo {author} {\bibfnamefont {G.~B.}\ \bibnamefont {Arous}}, \ and\ \bibinfo {author} {\bibfnamefont {G.}~\bibnamefont {Biroli}},\ }in\ \href@noop {} {\emph {\bibinfo {booktitle} {Mathematical and Scientific Machine Learning}}}\ (\bibinfo {organization} {PMLR},\ \bibinfo {year} {2020})\ pp.\ \bibinfo {pages} {287--327}\BibitemShut {NoStop}%
\bibitem [{\citenamefont {Arjevani}\ and\ \citenamefont {Field}(2020)}]{Arjevani_Field_2020}%
  \BibitemOpen
  \bibfield  {author} {\bibinfo {author} {\bibfnamefont {Y.}~\bibnamefont {Arjevani}}\ and\ \bibinfo {author} {\bibfnamefont {M.}~\bibnamefont {Field}},\ }in\ \href {https://proceedings.neurips.cc/paper/2020/hash/3a61ed715ee66c48bacf237fa7bb5289-Abstract.html} {\emph {\bibinfo {booktitle} {Advances in Neural Information Processing Systems}}},\ Vol.~\bibinfo {volume} {33}\ (\bibinfo  {publisher} {Curran Associates, Inc.},\ \bibinfo {year} {2020})\ p.\ \bibinfo {pages} {5441–5452}\BibitemShut {NoStop}%
\bibitem [{\citenamefont {Arjevani}\ and\ \citenamefont {Field}(2021)}]{Arjevani_Field_2021}%
  \BibitemOpen
  \bibfield  {author} {\bibinfo {author} {\bibfnamefont {Y.}~\bibnamefont {Arjevani}}\ and\ \bibinfo {author} {\bibfnamefont {M.}~\bibnamefont {Field}},\ }in\ \href {https://proceedings.neurips.cc/paper/2021/hash/806d926414ce19d907700e23177ab4ff-Abstract.html} {\emph {\bibinfo {booktitle} {Advances in Neural Information Processing Systems}}},\ Vol.~\bibinfo {volume} {34}\ (\bibinfo  {publisher} {Curran Associates, Inc.},\ \bibinfo {year} {2021})\ p.\ \bibinfo {pages} {15162–15174}\BibitemShut {NoStop}%
\bibitem [{\citenamefont {Fyodorov}\ and\ \citenamefont {Tublin}(2022)}]{fyodorov2022optimization}%
  \BibitemOpen
  \bibfield  {author} {\bibinfo {author} {\bibfnamefont {Y.~V.}\ \bibnamefont {Fyodorov}}\ and\ \bibinfo {author} {\bibfnamefont {R.}~\bibnamefont {Tublin}},\ }\href@noop {} {\bibfield  {journal} {\bibinfo  {journal} {Journal of Physics A: Mathematical and Theoretical}\ }\textbf {\bibinfo {volume} {55}},\ \bibinfo {pages} {244008} (\bibinfo {year} {2022})}\BibitemShut {NoStop}%
\bibitem [{\citenamefont {Mcculloch}\ and\ \citenamefont {Pitts}(1943)}]{mcculloch43a}%
  \BibitemOpen
  \bibfield  {author} {\bibinfo {author} {\bibfnamefont {W.}~\bibnamefont {Mcculloch}}\ and\ \bibinfo {author} {\bibfnamefont {W.}~\bibnamefont {Pitts}},\ }\href@noop {} {\bibfield  {journal} {\bibinfo  {journal} {Bulletin of Mathematical Biophysics}\ }\textbf {\bibinfo {volume} {5}},\ \bibinfo {pages} {127} (\bibinfo {year} {1943})}\BibitemShut {NoStop}%
\bibitem [{\citenamefont {Derrida}\ \emph {et~al.}(1987)\citenamefont {Derrida}, \citenamefont {Gardner},\ and\ \citenamefont {Zippelius}}]{Derrida_Gardner_Zippelius_1987}%
  \BibitemOpen
  \bibfield  {author} {\bibinfo {author} {\bibfnamefont {B.}~\bibnamefont {Derrida}}, \bibinfo {author} {\bibfnamefont {E.}~\bibnamefont {Gardner}}, \ and\ \bibinfo {author} {\bibfnamefont {A.}~\bibnamefont {Zippelius}},\ }\href {\doibase 10.1209/0295-5075/4/2/007} {\bibfield  {journal} {\bibinfo  {journal} {Europhysics Letters}\ }\textbf {\bibinfo {volume} {4}},\ \bibinfo {pages} {167} (\bibinfo {year} {1987})}\BibitemShut {NoStop}%
\bibitem [{\citenamefont {Gardner}\ and\ \citenamefont {Derrida}(1988)}]{Gardner_Derrida_1988}%
  \BibitemOpen
  \bibfield  {author} {\bibinfo {author} {\bibfnamefont {E.}~\bibnamefont {Gardner}}\ and\ \bibinfo {author} {\bibfnamefont {B.}~\bibnamefont {Derrida}},\ }\href {\doibase 10.1088/0305-4470/21/1/031} {\bibfield  {journal} {\bibinfo  {journal} {Journal of Physics A: Mathematical and General}\ }\textbf {\bibinfo {volume} {21}},\ \bibinfo {pages} {271} (\bibinfo {year} {1988})}\BibitemShut {NoStop}%
\bibitem [{\citenamefont {Gardner}\ and\ \citenamefont {Derrida}(1989)}]{Gardner_Derrida_1989}%
  \BibitemOpen
  \bibfield  {author} {\bibinfo {author} {\bibfnamefont {E.}~\bibnamefont {Gardner}}\ and\ \bibinfo {author} {\bibfnamefont {B.}~\bibnamefont {Derrida}},\ }\href {\doibase 10.1088/0305-4470/22/12/004} {\bibfield  {journal} {\bibinfo  {journal} {Journal of Physics A: Mathematical and General}\ }\textbf {\bibinfo {volume} {22}},\ \bibinfo {pages} {1983} (\bibinfo {year} {1989})}\BibitemShut {NoStop}%
\bibitem [{\citenamefont {Franz}\ and\ \citenamefont {Parisi}(2016)}]{franz2016simplest}%
  \BibitemOpen
  \bibfield  {author} {\bibinfo {author} {\bibfnamefont {S.}~\bibnamefont {Franz}}\ and\ \bibinfo {author} {\bibfnamefont {G.}~\bibnamefont {Parisi}},\ }\href@noop {} {\bibfield  {journal} {\bibinfo  {journal} {Journal of Physics A: Mathematical and Theoretical}\ }\textbf {\bibinfo {volume} {49}},\ \bibinfo {pages} {145001} (\bibinfo {year} {2016})}\BibitemShut {NoStop}%
\bibitem [{\citenamefont {Aza{\"\i}s}\ and\ \citenamefont {Wschebor}(2009)}]{azais2009level}%
  \BibitemOpen
  \bibfield  {author} {\bibinfo {author} {\bibfnamefont {J.-M.}\ \bibnamefont {Aza{\"\i}s}}\ and\ \bibinfo {author} {\bibfnamefont {M.}~\bibnamefont {Wschebor}},\ }\href@noop {} {\emph {\bibinfo {title} {Level sets and extrema of random processes and fields}}}\ (\bibinfo  {publisher} {John Wiley \& Sons},\ \bibinfo {year} {2009})\BibitemShut {NoStop}%
\bibitem [{\citenamefont {Auffinger}(2011)}]{Auffinger}%
  \BibitemOpen
  \bibfield  {author} {\bibinfo {author} {\bibfnamefont {A.}~\bibnamefont {Auffinger}},\ }\emph {\bibinfo {title} {Random Matrices, Complexity of Spin Glasses and Heavy Tailed Processes}},\ \href {https://www.proquest.com/docview/883392368/abstract/51E7E4C7E8964A09PQ/1} {\bibinfo {type} {Ph.d.}},\ \bibinfo  {school} {New York University}, \bibinfo {address} {United States -- New York} (\bibinfo {year} {2011})\BibitemShut {NoStop}%
\bibitem [{\citenamefont {Ros}\ and\ \citenamefont {Fyodorov}(2022)}]{Ros_Fyodorov_2023}%
  \BibitemOpen
  \bibfield  {author} {\bibinfo {author} {\bibfnamefont {V.}~\bibnamefont {Ros}}\ and\ \bibinfo {author} {\bibfnamefont {Y.~V.}\ \bibnamefont {Fyodorov}},\ }\href {http://arxiv.org/abs/2209.07975} {\bibfield  {journal} {\bibinfo  {journal} {Spin Glass Theory and Far Beyond: Replica Symmetry Breaking after 40 Years}\ ,\ \bibinfo {pages} {24}} (\bibinfo {year} {2022})}\BibitemShut {NoStop}%
\bibitem [{\citenamefont {Asgari}\ \emph {et~al.}(2025)\citenamefont {Asgari}, \citenamefont {Montanari},\ and\ \citenamefont {Saeed}}]{asgari2025localminimaempiricalrisk}%
  \BibitemOpen
  \bibfield  {author} {\bibinfo {author} {\bibfnamefont {K.}~\bibnamefont {Asgari}}, \bibinfo {author} {\bibfnamefont {A.}~\bibnamefont {Montanari}}, \ and\ \bibinfo {author} {\bibfnamefont {B.}~\bibnamefont {Saeed}},\ }\href {https://arxiv.org/abs/2502.01953} {\enquote {\bibinfo {title} {Local minima of the empirical risk in high dimension: General theorems and convex examples},}\ } (\bibinfo {year} {2025}),\ \Eprint {http://arxiv.org/abs/2502.01953} {arXiv:2502.01953 [stat.ML]} \BibitemShut {NoStop}%
\bibitem [{\citenamefont {Kingma}(2013)}]{kingma2013auto}%
  \BibitemOpen
  \bibfield  {author} {\bibinfo {author} {\bibfnamefont {D.~P.}\ \bibnamefont {Kingma}},\ }\href@noop {} {\bibfield  {journal} {\bibinfo  {journal} {arXiv preprint arXiv:1312.6114}\ } (\bibinfo {year} {2013})}\BibitemShut {NoStop}%
\bibitem [{\citenamefont {Goodfellow}\ \emph {et~al.}(2014)\citenamefont {Goodfellow}, \citenamefont {Pouget-Abadie}, \citenamefont {Mirza}, \citenamefont {Xu}, \citenamefont {Warde-Farley}, \citenamefont {Ozair}, \citenamefont {Courville},\ and\ \citenamefont {Bengio}}]{goodfellow2014generative}%
  \BibitemOpen
  \bibfield  {author} {\bibinfo {author} {\bibfnamefont {I.}~\bibnamefont {Goodfellow}}, \bibinfo {author} {\bibfnamefont {J.}~\bibnamefont {Pouget-Abadie}}, \bibinfo {author} {\bibfnamefont {M.}~\bibnamefont {Mirza}}, \bibinfo {author} {\bibfnamefont {B.}~\bibnamefont {Xu}}, \bibinfo {author} {\bibfnamefont {D.}~\bibnamefont {Warde-Farley}}, \bibinfo {author} {\bibfnamefont {S.}~\bibnamefont {Ozair}}, \bibinfo {author} {\bibfnamefont {A.}~\bibnamefont {Courville}}, \ and\ \bibinfo {author} {\bibfnamefont {Y.}~\bibnamefont {Bengio}},\ }\href@noop {} {\bibfield  {journal} {\bibinfo  {journal} {Advances in neural information processing systems}\ }\textbf {\bibinfo {volume} {27}} (\bibinfo {year} {2014})}\BibitemShut {NoStop}%
\bibitem [{\citenamefont {Seddik}\ \emph {et~al.}(2020)\citenamefont {Seddik}, \citenamefont {Louart}, \citenamefont {Tamaazousti},\ and\ \citenamefont {Couillet}}]{seddik2020random}%
  \BibitemOpen
  \bibfield  {author} {\bibinfo {author} {\bibfnamefont {M.~E.~A.}\ \bibnamefont {Seddik}}, \bibinfo {author} {\bibfnamefont {C.}~\bibnamefont {Louart}}, \bibinfo {author} {\bibfnamefont {M.}~\bibnamefont {Tamaazousti}}, \ and\ \bibinfo {author} {\bibfnamefont {R.}~\bibnamefont {Couillet}},\ }in\ \href@noop {} {\emph {\bibinfo {booktitle} {International Conference on Machine Learning}}}\ (\bibinfo {organization} {PMLR},\ \bibinfo {year} {2020})\ pp.\ \bibinfo {pages} {8573--8582}\BibitemShut {NoStop}%
\bibitem [{\citenamefont {Boucheron}\ \emph {et~al.}(2003)\citenamefont {Boucheron}, \citenamefont {Lugosi},\ and\ \citenamefont {Bousquet}}]{boucheron2003concentration}%
  \BibitemOpen
  \bibfield  {author} {\bibinfo {author} {\bibfnamefont {S.}~\bibnamefont {Boucheron}}, \bibinfo {author} {\bibfnamefont {G.}~\bibnamefont {Lugosi}}, \ and\ \bibinfo {author} {\bibfnamefont {O.}~\bibnamefont {Bousquet}},\ }in\ \href@noop {} {\emph {\bibinfo {booktitle} {Summer school on machine learning}}}\ (\bibinfo  {publisher} {Springer},\ \bibinfo {year} {2003})\ pp.\ \bibinfo {pages} {208--240}\BibitemShut {NoStop}%
\bibitem [{\citenamefont {Wainwright}(2019)}]{wainwright2019high}%
  \BibitemOpen
  \bibfield  {author} {\bibinfo {author} {\bibfnamefont {M.~J.}\ \bibnamefont {Wainwright}},\ }\href@noop {} {\emph {\bibinfo {title} {High-dimensional statistics: A non-asymptotic viewpoint}}},\ Vol.~\bibinfo {volume} {48}\ (\bibinfo  {publisher} {Cambridge university press},\ \bibinfo {year} {2019})\BibitemShut {NoStop}%
\bibitem [{\citenamefont {Sohl-Dickstein}\ \emph {et~al.}(2015)\citenamefont {Sohl-Dickstein}, \citenamefont {Weiss}, \citenamefont {Maheswaranathan},\ and\ \citenamefont {Ganguli}}]{Sohl-Dickstein_Weiss_Maheswaranathan_Ganguli_2015}%
  \BibitemOpen
  \bibfield  {author} {\bibinfo {author} {\bibfnamefont {J.}~\bibnamefont {Sohl-Dickstein}}, \bibinfo {author} {\bibfnamefont {E.}~\bibnamefont {Weiss}}, \bibinfo {author} {\bibfnamefont {N.}~\bibnamefont {Maheswaranathan}}, \ and\ \bibinfo {author} {\bibfnamefont {S.}~\bibnamefont {Ganguli}},\ }in\ \href@noop {} {\emph {\bibinfo {booktitle} {International conference on machine learning}}}\ (\bibinfo {organization} {pmlr},\ \bibinfo {year} {2015})\ pp.\ \bibinfo {pages} {2256--2265}\BibitemShut {NoStop}%
\bibitem [{\citenamefont {Song}\ and\ \citenamefont {Ermon}(2019)}]{Song_Ermon_2019}%
  \BibitemOpen
  \bibfield  {author} {\bibinfo {author} {\bibfnamefont {Y.}~\bibnamefont {Song}}\ and\ \bibinfo {author} {\bibfnamefont {S.}~\bibnamefont {Ermon}},\ }in\ \href {https://proceedings.neurips.cc/paper_files/paper/2019/hash/3001ef257407d5a371a96dcd947c7d93-Abstract.html} {\emph {\bibinfo {booktitle} {Advances in Neural Information Processing Systems}}},\ Vol.~\bibinfo {volume} {32}\ (\bibinfo  {publisher} {Curran Associates, Inc.},\ \bibinfo {year} {2019})\BibitemShut {NoStop}%
\bibitem [{\citenamefont {Ho}\ \emph {et~al.}(2020)\citenamefont {Ho}, \citenamefont {Jain},\ and\ \citenamefont {Abbeel}}]{Ho_Jain_Abbeel_2020}%
  \BibitemOpen
  \bibfield  {author} {\bibinfo {author} {\bibfnamefont {J.}~\bibnamefont {Ho}}, \bibinfo {author} {\bibfnamefont {A.}~\bibnamefont {Jain}}, \ and\ \bibinfo {author} {\bibfnamefont {P.}~\bibnamefont {Abbeel}},\ }\href@noop {} {\bibfield  {journal} {\bibinfo  {journal} {Advances in neural information processing systems}\ }\textbf {\bibinfo {volume} {33}},\ \bibinfo {pages} {6840} (\bibinfo {year} {2020})}\BibitemShut {NoStop}%
\bibitem [{\citenamefont {Cauchy}\ \emph {et~al.}(1847)\citenamefont {Cauchy} \emph {et~al.}}]{cauchy1847methode}%
  \BibitemOpen
  \bibfield  {author} {\bibinfo {author} {\bibfnamefont {A.}~\bibnamefont {Cauchy}} \emph {et~al.},\ }\href@noop {} {\bibfield  {journal} {\bibinfo  {journal} {Comp. Rend. Sci. Paris}\ }\textbf {\bibinfo {volume} {25}},\ \bibinfo {pages} {536} (\bibinfo {year} {1847})}\BibitemShut {NoStop}%
\bibitem [{\citenamefont {Robbins}\ and\ \citenamefont {Siegmund}(1971)}]{robbins1971convergence}%
  \BibitemOpen
  \bibfield  {author} {\bibinfo {author} {\bibfnamefont {H.}~\bibnamefont {Robbins}}\ and\ \bibinfo {author} {\bibfnamefont {D.}~\bibnamefont {Siegmund}},\ }in\ \href@noop {} {\emph {\bibinfo {booktitle} {Optimizing methods in statistics}}}\ (\bibinfo  {publisher} {Elsevier},\ \bibinfo {year} {1971})\ pp.\ \bibinfo {pages} {233--257}\BibitemShut {NoStop}%
\bibitem [{\citenamefont {Bena{\"\i}m}(2006)}]{benaim2006dynamics}%
  \BibitemOpen
  \bibfield  {author} {\bibinfo {author} {\bibfnamefont {M.}~\bibnamefont {Bena{\"\i}m}},\ }in\ \href@noop {} {\emph {\bibinfo {booktitle} {Seminaire de probabilites XXXIII}}}\ (\bibinfo  {publisher} {Springer},\ \bibinfo {year} {2006})\ pp.\ \bibinfo {pages} {1--68}\BibitemShut {NoStop}%
\bibitem [{\citenamefont {Panageas}\ and\ \citenamefont {Piliouras}(2016)}]{panageas2016gradient}%
  \BibitemOpen
  \bibfield  {author} {\bibinfo {author} {\bibfnamefont {I.}~\bibnamefont {Panageas}}\ and\ \bibinfo {author} {\bibfnamefont {G.}~\bibnamefont {Piliouras}},\ }\href@noop {} {\bibfield  {journal} {\bibinfo  {journal} {arXiv preprint arXiv:1605.00405}\ } (\bibinfo {year} {2016})}\BibitemShut {NoStop}%
\bibitem [{\citenamefont {Panageas}\ \emph {et~al.}(2019)\citenamefont {Panageas}, \citenamefont {Piliouras},\ and\ \citenamefont {Wang}}]{panageas2019first}%
  \BibitemOpen
  \bibfield  {author} {\bibinfo {author} {\bibfnamefont {I.}~\bibnamefont {Panageas}}, \bibinfo {author} {\bibfnamefont {G.}~\bibnamefont {Piliouras}}, \ and\ \bibinfo {author} {\bibfnamefont {X.}~\bibnamefont {Wang}},\ }\href@noop {} {\bibfield  {journal} {\bibinfo  {journal} {Advances in Neural Information Processing Systems}\ }\textbf {\bibinfo {volume} {32}} (\bibinfo {year} {2019})}\BibitemShut {NoStop}%
\bibitem [{\citenamefont {Antonakopoulos}\ \emph {et~al.}(2022)\citenamefont {Antonakopoulos}, \citenamefont {Mertikopoulos}, \citenamefont {Piliouras},\ and\ \citenamefont {Wang}}]{antonakopoulos2022adagrad}%
  \BibitemOpen
  \bibfield  {author} {\bibinfo {author} {\bibfnamefont {K.}~\bibnamefont {Antonakopoulos}}, \bibinfo {author} {\bibfnamefont {P.}~\bibnamefont {Mertikopoulos}}, \bibinfo {author} {\bibfnamefont {G.}~\bibnamefont {Piliouras}}, \ and\ \bibinfo {author} {\bibfnamefont {X.}~\bibnamefont {Wang}},\ }in\ \href@noop {} {\emph {\bibinfo {booktitle} {International Conference on Machine Learning}}}\ (\bibinfo {organization} {PMLR},\ \bibinfo {year} {2022})\ pp.\ \bibinfo {pages} {731--771}\BibitemShut {NoStop}%
\bibitem [{\citenamefont {Saad}\ and\ \citenamefont {Solla}(1995)}]{saad1995exact}%
  \BibitemOpen
  \bibfield  {author} {\bibinfo {author} {\bibfnamefont {D.}~\bibnamefont {Saad}}\ and\ \bibinfo {author} {\bibfnamefont {S.~A.}\ \bibnamefont {Solla}},\ }\href@noop {} {\bibfield  {journal} {\bibinfo  {journal} {Physical Review Letters}\ }\textbf {\bibinfo {volume} {74}},\ \bibinfo {pages} {4337} (\bibinfo {year} {1995})}\BibitemShut {NoStop}%
\bibitem [{\citenamefont {Veiga}\ \emph {et~al.}(2022)\citenamefont {Veiga}, \citenamefont {Stephan}, \citenamefont {Loureiro}, \citenamefont {Krzakala},\ and\ \citenamefont {Zdeborov{\'a}}}]{veiga2022phase}%
  \BibitemOpen
  \bibfield  {author} {\bibinfo {author} {\bibfnamefont {R.}~\bibnamefont {Veiga}}, \bibinfo {author} {\bibfnamefont {L.}~\bibnamefont {Stephan}}, \bibinfo {author} {\bibfnamefont {B.}~\bibnamefont {Loureiro}}, \bibinfo {author} {\bibfnamefont {F.}~\bibnamefont {Krzakala}}, \ and\ \bibinfo {author} {\bibfnamefont {L.}~\bibnamefont {Zdeborov{\'a}}},\ }\href@noop {} {\bibfield  {journal} {\bibinfo  {journal} {Advances in Neural Information Processing Systems}\ }\textbf {\bibinfo {volume} {35}},\ \bibinfo {pages} {23244} (\bibinfo {year} {2022})}\BibitemShut {NoStop}%
\bibitem [{\citenamefont {Fyodorov}(2004)}]{fyodorov2004complexity}%
  \BibitemOpen
  \bibfield  {author} {\bibinfo {author} {\bibfnamefont {Y.~V.}\ \bibnamefont {Fyodorov}},\ }\href@noop {} {\bibfield  {journal} {\bibinfo  {journal} {Physical review letters}\ }\textbf {\bibinfo {volume} {92}},\ \bibinfo {pages} {240601} (\bibinfo {year} {2004})}\BibitemShut {NoStop}%
\bibitem [{\citenamefont {Fyodorov}(2005)}]{fyodorov2005counting}%
  \BibitemOpen
  \bibfield  {author} {\bibinfo {author} {\bibfnamefont {Y.~V.}\ \bibnamefont {Fyodorov}},\ }\href@noop {} {\bibfield  {journal} {\bibinfo  {journal} {arXiv preprint cond-mat/0507059}\ } (\bibinfo {year} {2005})}\BibitemShut {NoStop}%
\bibitem [{\citenamefont {Fyodorov}\ and\ \citenamefont {Williams}(2007)}]{fyodorov2007replica}%
  \BibitemOpen
  \bibfield  {author} {\bibinfo {author} {\bibfnamefont {Y.~V.}\ \bibnamefont {Fyodorov}}\ and\ \bibinfo {author} {\bibfnamefont {I.}~\bibnamefont {Williams}},\ }\href@noop {} {\bibfield  {journal} {\bibinfo  {journal} {Journal of Statistical Physics}\ }\textbf {\bibinfo {volume} {129}},\ \bibinfo {pages} {1081} (\bibinfo {year} {2007})}\BibitemShut {NoStop}%
\bibitem [{\citenamefont {Fyodorov}\ and\ \citenamefont {Keating}(2014)}]{fyodorov2014freezing}%
  \BibitemOpen
  \bibfield  {author} {\bibinfo {author} {\bibfnamefont {Y.~V.}\ \bibnamefont {Fyodorov}}\ and\ \bibinfo {author} {\bibfnamefont {J.~P.}\ \bibnamefont {Keating}},\ }\href@noop {} {\bibfield  {journal} {\bibinfo  {journal} {Philosophical Transactions of the Royal Society A: Mathematical, Physical and Engineering Sciences}\ }\textbf {\bibinfo {volume} {372}},\ \bibinfo {pages} {20120503} (\bibinfo {year} {2014})}\BibitemShut {NoStop}%
\bibitem [{\citenamefont {Ben~Arous}\ \emph {et~al.}(2023)\citenamefont {Ben~Arous}, \citenamefont {Bourgade},\ and\ \citenamefont {McKenna}}]{Arous_Bourgade_McKenna_2022}%
  \BibitemOpen
  \bibfield  {author} {\bibinfo {author} {\bibfnamefont {G.}~\bibnamefont {Ben~Arous}}, \bibinfo {author} {\bibfnamefont {P.}~\bibnamefont {Bourgade}}, \ and\ \bibinfo {author} {\bibfnamefont {B.}~\bibnamefont {McKenna}},\ }\href@noop {} {\bibfield  {journal} {\bibinfo  {journal} {Probability and Mathematical Physics}\ }\textbf {\bibinfo {volume} {3}},\ \bibinfo {pages} {731} (\bibinfo {year} {2023})}\BibitemShut {NoStop}%
\bibitem [{\citenamefont {Auffinger}\ and\ \citenamefont {Chen}(2014)}]{Auffinger_Chen_2014}%
  \BibitemOpen
  \bibfield  {author} {\bibinfo {author} {\bibfnamefont {A.}~\bibnamefont {Auffinger}}\ and\ \bibinfo {author} {\bibfnamefont {W.-K.}\ \bibnamefont {Chen}},\ }\href {\doibase 10.1007/s10955-014-1073-0} {\bibfield  {journal} {\bibinfo  {journal} {Journal of Statistical Physics}\ }\textbf {\bibinfo {volume} {157}},\ \bibinfo {pages} {40–59} (\bibinfo {year} {2014})}\BibitemShut {NoStop}%
\bibitem [{\citenamefont {Fan}\ \emph {et~al.}(2021)\citenamefont {Fan}, \citenamefont {Mei},\ and\ \citenamefont {Montanari}}]{Fan_Mei_Montanari_2020}%
  \BibitemOpen
  \bibfield  {author} {\bibinfo {author} {\bibfnamefont {Z.}~\bibnamefont {Fan}}, \bibinfo {author} {\bibfnamefont {S.}~\bibnamefont {Mei}}, \ and\ \bibinfo {author} {\bibfnamefont {A.}~\bibnamefont {Montanari}},\ }\href@noop {} {\bibfield  {journal} {\bibinfo  {journal} {The Annals of Probability}\ }\textbf {\bibinfo {volume} {49}},\ \bibinfo {pages} {1} (\bibinfo {year} {2021})}\BibitemShut {NoStop}%
\bibitem [{\citenamefont {McKenna}(2024)}]{McKenna_2022}%
  \BibitemOpen
  \bibfield  {author} {\bibinfo {author} {\bibfnamefont {B.}~\bibnamefont {McKenna}},\ }in\ \href@noop {} {\emph {\bibinfo {booktitle} {Annales de l'Institut Henri Poincare (B) Probabilites et statistiques}}},\ Vol.\ \bibinfo {volume} {60 (1)}\ (\bibinfo {organization} {Institut Henri Poincar{\'e}},\ \bibinfo {year} {2024})\ pp.\ \bibinfo {pages} {636--657}\BibitemShut {NoStop}%
\bibitem [{\citenamefont {Ros}\ \emph {et~al.}(2019)\citenamefont {Ros}, \citenamefont {Ben~Arous}, \citenamefont {Biroli},\ and\ \citenamefont {Cammarota}}]{Ros_Ben_Arous_Biroli_Cammarota_2019}%
  \BibitemOpen
  \bibfield  {author} {\bibinfo {author} {\bibfnamefont {V.}~\bibnamefont {Ros}}, \bibinfo {author} {\bibfnamefont {G.}~\bibnamefont {Ben~Arous}}, \bibinfo {author} {\bibfnamefont {G.}~\bibnamefont {Biroli}}, \ and\ \bibinfo {author} {\bibfnamefont {C.}~\bibnamefont {Cammarota}},\ }\href {\doibase 10.1103/PhysRevX.9.011003} {\bibfield  {journal} {\bibinfo  {journal} {Physical Review X}\ }\textbf {\bibinfo {volume} {9}},\ \bibinfo {pages} {011003} (\bibinfo {year} {2019})}\BibitemShut {NoStop}%
\bibitem [{\citenamefont {Zdeborová}\ and\ \citenamefont {Krzakala}(2016)}]{Zdeborová_Krzakala_2016}%
  \BibitemOpen
  \bibfield  {author} {\bibinfo {author} {\bibfnamefont {L.}~\bibnamefont {Zdeborová}}\ and\ \bibinfo {author} {\bibfnamefont {F.}~\bibnamefont {Krzakala}},\ }\href {\doibase 10.1080/00018732.2016.1211393} {\bibfield  {journal} {\bibinfo  {journal} {Advances in Physics}\ }\textbf {\bibinfo {volume} {65}},\ \bibinfo {pages} {453–552} (\bibinfo {year} {2016})}\BibitemShut {NoStop}%
\bibitem [{\citenamefont {Barbier}(2020)}]{Barbier_2020}%
  \BibitemOpen
  \bibfield  {author} {\bibinfo {author} {\bibfnamefont {J.}~\bibnamefont {Barbier}},\ }\href@noop {} {\bibfield  {journal} {\bibinfo  {journal} {arXiv preprint arXiv:2010.14863}\ } (\bibinfo {year} {2020})}\BibitemShut {NoStop}%
\bibitem [{\citenamefont {Gamarnik}\ \emph {et~al.}(2022)\citenamefont {Gamarnik}, \citenamefont {Moore},\ and\ \citenamefont {Zdeborov{\'a}}}]{Gamarnik_Moore_Zdeborová_2022}%
  \BibitemOpen
  \bibfield  {author} {\bibinfo {author} {\bibfnamefont {D.}~\bibnamefont {Gamarnik}}, \bibinfo {author} {\bibfnamefont {C.}~\bibnamefont {Moore}}, \ and\ \bibinfo {author} {\bibfnamefont {L.}~\bibnamefont {Zdeborov{\'a}}},\ }\href@noop {} {\bibfield  {journal} {\bibinfo  {journal} {Journal of Statistical Mechanics: Theory and Experiment}\ }\textbf {\bibinfo {volume} {2022}},\ \bibinfo {pages} {114015} (\bibinfo {year} {2022})}\BibitemShut {NoStop}%
\bibitem [{\citenamefont {Choromanska}\ \emph {et~al.}(2015{\natexlab{b}})\citenamefont {Choromanska}, \citenamefont {LeCun},\ and\ \citenamefont {Arous}}]{Choromanska_LeCun_Arous_2015}%
  \BibitemOpen
  \bibfield  {author} {\bibinfo {author} {\bibfnamefont {A.}~\bibnamefont {Choromanska}}, \bibinfo {author} {\bibfnamefont {Y.}~\bibnamefont {LeCun}}, \ and\ \bibinfo {author} {\bibfnamefont {G.~B.}\ \bibnamefont {Arous}},\ }in\ \href {https://proceedings.mlr.press/v40/Choromanska15.html} {\emph {\bibinfo {booktitle} {Proceedings of The 28th Conference on Learning Theory}}}\ (\bibinfo  {publisher} {PMLR},\ \bibinfo {year} {2015})\ p.\ \bibinfo {pages} {1756–1760}\BibitemShut {NoStop}%
\bibitem [{\citenamefont {Hartnett}\ \emph {et~al.}(2018)\citenamefont {Hartnett}, \citenamefont {Parker},\ and\ \citenamefont {Geist}}]{Hartnett_Parker_Geist_2018}%
  \BibitemOpen
  \bibfield  {author} {\bibinfo {author} {\bibfnamefont {G.~S.}\ \bibnamefont {Hartnett}}, \bibinfo {author} {\bibfnamefont {E.}~\bibnamefont {Parker}}, \ and\ \bibinfo {author} {\bibfnamefont {E.}~\bibnamefont {Geist}},\ }\href {\doibase 10.1103/PhysRevE.98.022116} {\bibfield  {journal} {\bibinfo  {journal} {Physical Review E}\ }\textbf {\bibinfo {volume} {98}},\ \bibinfo {pages} {022116} (\bibinfo {year} {2018})}\BibitemShut {NoStop}%
\bibitem [{\citenamefont {Moustakas}\ \emph {et~al.}(2003)\citenamefont {Moustakas}, \citenamefont {Simon},\ and\ \citenamefont {Sengupta}}]{Moustakas2003MIMO1}%
  \BibitemOpen
  \bibfield  {author} {\bibinfo {author} {\bibfnamefont {A.~L.}\ \bibnamefont {Moustakas}}, \bibinfo {author} {\bibfnamefont {S.~H.}\ \bibnamefont {Simon}}, \ and\ \bibinfo {author} {\bibfnamefont {A.~M.}\ \bibnamefont {Sengupta}},\ }\href@noop {} {\bibfield  {journal} {\bibinfo  {journal} {{IEEE} Trans. on Information Theory}\ }\textbf {\bibinfo {volume} {49}},\ \bibinfo {pages} {2545} (\bibinfo {year} {2003})}\BibitemShut {NoStop}%
\bibitem [{\citenamefont {Hachem}\ \emph {et~al.}(2008)\citenamefont {Hachem}, \citenamefont {Khorunzhiy}, \citenamefont {Loubaton}, \citenamefont {Najim},\ and\ \citenamefont {Pastur}}]{Hachem2006_GaussianCapacityKroneckerProduct}%
  \BibitemOpen
  \bibfield  {author} {\bibinfo {author} {\bibfnamefont {W.}~\bibnamefont {Hachem}}, \bibinfo {author} {\bibfnamefont {O.}~\bibnamefont {Khorunzhiy}}, \bibinfo {author} {\bibfnamefont {P.}~\bibnamefont {Loubaton}}, \bibinfo {author} {\bibfnamefont {J.}~\bibnamefont {Najim}}, \ and\ \bibinfo {author} {\bibfnamefont {L.}~\bibnamefont {Pastur}},\ }\href@noop {} {\bibfield  {journal} {\bibinfo  {journal} {{IEEE} Trans. Inform. Theory}\ }\textbf {\bibinfo {volume} {54}},\ \bibinfo {pages} {3987} (\bibinfo {year} {2008})}\BibitemShut {NoStop}%
\bibitem [{\citenamefont {Moustakas}\ and\ \citenamefont {Kazakopoulos}(2013)}]{moustakas2013sinr}%
  \BibitemOpen
  \bibfield  {author} {\bibinfo {author} {\bibfnamefont {A.~L.}\ \bibnamefont {Moustakas}}\ and\ \bibinfo {author} {\bibfnamefont {P.}~\bibnamefont {Kazakopoulos}},\ }\href@noop {} {\bibfield  {journal} {\bibinfo  {journal} {{IEEE} Trans. Inform. Theory}\ }\textbf {\bibinfo {volume} {59}},\ \bibinfo {pages} {6490} (\bibinfo {year} {2013})}\BibitemShut {NoStop}%
\bibitem [{\citenamefont {Paul}\ and\ \citenamefont {Silverstein}(2009)}]{paul2009no}%
  \BibitemOpen
  \bibfield  {author} {\bibinfo {author} {\bibfnamefont {D.}~\bibnamefont {Paul}}\ and\ \bibinfo {author} {\bibfnamefont {J.~W.}\ \bibnamefont {Silverstein}},\ }\href@noop {} {\bibfield  {journal} {\bibinfo  {journal} {Journal of Multivariate Analysis}\ }\textbf {\bibinfo {volume} {100}},\ \bibinfo {pages} {37} (\bibinfo {year} {2009})}\BibitemShut {NoStop}%
\bibitem [{\citenamefont {Bercu}\ \emph {et~al.}(2000)\citenamefont {Bercu}, \citenamefont {Gamboa},\ and\ \citenamefont {Lavielle}}]{bercu2000sharp}%
  \BibitemOpen
  \bibfield  {author} {\bibinfo {author} {\bibfnamefont {B.}~\bibnamefont {Bercu}}, \bibinfo {author} {\bibfnamefont {F.}~\bibnamefont {Gamboa}}, \ and\ \bibinfo {author} {\bibfnamefont {M.}~\bibnamefont {Lavielle}},\ }\href@noop {} {\bibfield  {journal} {\bibinfo  {journal} {ESAIM: Probability and Statistics}\ }\textbf {\bibinfo {volume} {4}},\ \bibinfo {pages} {1} (\bibinfo {year} {2000})}\BibitemShut {NoStop}%
\bibitem [{\citenamefont {Dembo}\ and\ \citenamefont {Zeitouni}(2009)}]{dembo2009large}%
  \BibitemOpen
  \bibfield  {author} {\bibinfo {author} {\bibfnamefont {A.}~\bibnamefont {Dembo}}\ and\ \bibinfo {author} {\bibfnamefont {O.}~\bibnamefont {Zeitouni}},\ }\href@noop {} {\emph {\bibinfo {title} {Large deviations techniques and applications}}},\ Vol.~\bibinfo {volume} {38}\ (\bibinfo  {publisher} {Springer Science \& Business Media},\ \bibinfo {year} {2009})\BibitemShut {NoStop}%
\bibitem [{\citenamefont {Moustakas}\ and\ \citenamefont {Simon}(2007)}]{Moustakas2007_MIMO1}%
  \BibitemOpen
  \bibfield  {author} {\bibinfo {author} {\bibfnamefont {A.~L.}\ \bibnamefont {Moustakas}}\ and\ \bibinfo {author} {\bibfnamefont {S.~H.}\ \bibnamefont {Simon}},\ }\href@noop {} {\bibfield  {journal} {\bibinfo  {journal} {{IEEE} Trans. Inform. Theory}\ }\textbf {\bibinfo {volume} {53}},\ \bibinfo {pages} {3887} (\bibinfo {year} {2007})}\BibitemShut {NoStop}%
\bibitem [{\citenamefont {Couillet}\ \emph {et~al.}(2011)\citenamefont {Couillet}, \citenamefont {Debbah},\ and\ \citenamefont {Silverstein}}]{Couillet_Debbah_Silverstein_2011}%
  \BibitemOpen
  \bibfield  {author} {\bibinfo {author} {\bibfnamefont {R.}~\bibnamefont {Couillet}}, \bibinfo {author} {\bibfnamefont {M.}~\bibnamefont {Debbah}}, \ and\ \bibinfo {author} {\bibfnamefont {J.~W.}\ \bibnamefont {Silverstein}},\ }\href {\doibase 10.1109/TIT.2011.2133151} {\bibfield  {journal} {\bibinfo  {journal} {{IEEE} Trans. Inform. Theory}\ }\textbf {\bibinfo {volume} {57}},\ \bibinfo {pages} {3493–3514} (\bibinfo {year} {2011})}\BibitemShut {NoStop}%
\end{thebibliography}
\bibliographystyle{apsrev4-1}

\end{document}